\documentclass[11pt]{article}
\AtBeginDocument{%
  }

\usepackage[a4paper,margin=1in]{geometry}
\usepackage[T1]{fontenc}
\usepackage[utf8]{inputenc}
\usepackage{lmodern}
\usepackage{amsmath,amsfonts}
\usepackage{graphicx}
\usepackage{array}
\usepackage{tabularx}
\usepackage{multirow}
\usepackage{booktabs}
\usepackage{makecell}
\usepackage{placeins}
\usepackage[authoryear,round]{natbib}
\usepackage{microtype}
\usepackage{xcolor}
\usepackage{hyperref}

\hypersetup{
  colorlinks=true,
  linkcolor=blue!55!black,
  citecolor=blue!55!black,
  urlcolor=blue!55!black,
  pdftitle={Reflex-Informed Neuromuscular Reinforcement Learning for Muscle-Driven Locomotion},
  pdfauthor={Jian Zhou; Xingyu Zhang; Rui Ma; Yu Cao; Shane Xie; Zhi-qiang Zhang}
}

\newcommand{\suppvideo}{supplementary video}
\newcolumntype{L}[1]{>{\raggedright\arraybackslash}m{#1}}
\newcolumntype{Y}{>{\raggedright\arraybackslash}X}
\providecommand{\Description}[1]{}
\providecommand{\keywords}[1]{%
  \par\vspace{0.75em}\noindent\textbf{Keywords:} #1\par}
\begin{document}

\title{
Reflex-Informed Neuromuscular Reinforcement Learning for Muscle-Driven Locomotion
}
\author{Jian Zhou \quad Xingyu Zhang \quad Rui Ma\\[0.35em]
Yu Cao\textsuperscript{*} \quad Shane Xie \quad
Zhi-qiang Zhang\textsuperscript{*}\\[0.55em]
\small University of Leeds, Leeds, United Kingdom\\[0.25em]
\footnotesize \texttt{eljzho@leeds.ac.uk} \quad
\texttt{elxz@leeds.ac.uk} \quad \texttt{elrma@leeds.ac.uk}\\
\footnotesize \texttt{Y.Cao1@leeds.ac.uk} \quad
\texttt{S.Q.Xie@leeds.ac.uk} \quad \texttt{Z.Zhang3@leeds.ac.uk}\\[0.25em]
\small \textsuperscript{*}Corresponding authors}
\date{September 10, 2026}
\maketitle

\begin{abstract}

Muscle-driven locomotion provides a physically grounded approach to generating realistic human movement. However, achieving both physiological plausibility and adaptability to changes in musculoskeletal capacity and external disturbances remains a fundamental challenge. To address this limitation, we propose a Reflex-Informed Neuromuscular Reinforcement Learning framework for muscle-driven locomotion. Within this framework, a fixed phase-dependent reflex controller serves as the underlying neuromuscular control mechanism, while the reinforcement learning policy produces four biomechanically meaningful residual parameters to modulate key reflex gains and thresholds associated with hip swing, knee support, and ankle propulsion according to the current state. Experimental results demonstrate that the proposed framework generates physiologically plausible locomotion with improved kinematic accuracy and dynamic consistency, as well as better bilateral symmetry and stride-to-stride consistency under nominal walking conditions. The learned policy remains robust under muscle weakness and external perturbations without retraining.

\end{abstract}

\keywords{Muscle-driven locomotion, musculoskeletal simulation,
reflex-based control, residual reinforcement learning, neuromuscular control,
physics-based character animation}

\section{Introduction}
\label{sec:introduction}
Muscle-driven simulation plays an integral role in physically based character animation and the modeling of human locomotion \cite{sun2024,schumacher2025}. Unlike approaches that directly generate motion through joint trajectories or joint torques, muscle-driven models allow movement to emerge naturally from neuromuscular control, muscle dynamics, and physical interactions between the body and the environment, resulting in motion with greater physical consistency and biological plausibility. However, realistic muscle-driven locomotion requires not only physiological plausibility, but also the ability to adapt naturally to changes in the musculoskeletal condition and the environment. Therefore, achieving both physiological plausibility and adaptability remains a key challenge in muscle-driven character animation.

Existing approaches to muscle-driven locomotion generally fall into two categories. One category employs structured neuromuscular controllers, such as reflex-based control, central pattern generators (CPGs), and muscle synergies, by explicitly embedding biological control principles into the controller. Such controllers naturally generate stable and physiologically meaningful locomotion. However, they are typically designed manually or optimized offline, and their control policies remain fixed during deployment, limiting their ability to adapt to changes in musculoskeletal conditions or the environment.
The other category formulates locomotion as a reinforcement learning problem by directly learning muscle actions. Since muscle activations are generated online according to the current state, these methods exhibit strong adaptability. However, directly learning muscle actions requires the policy to discover effective neuromuscular coordination within a high-dimensional and redundant action space. Task-level rewards specify the desired locomotion objective but do not explicitly determine how muscle activity should be organized, making it difficult to ensure physiologically plausible movement and dynamics. Consequently, existing approaches often face a trade-off between physiological plausibility and adaptability: structured neuromuscular controllers preserve biological organization but lack state-dependent adaptation, whereas direct muscle learning remains adaptive but must rediscover complex neuromuscular coordination from scratch.


From the perspective of biological motor control, locomotion is not generated by independently controlling individual muscles. Instead, muscle activity is organized through existing neuromuscular mechanisms involving sensory feedback, spinal regulation, and muscle dynamics. In other words, existing neuromuscular mechanisms already organize how sensory feedback regulates muscle activity during locomotion. If these control principles already exist, should reinforcement learning still rediscover them, or should it instead build upon them?
This question motivates us to rethink the role of reinforcement learning in muscle-driven locomotion. Rather than directly learning muscle actions, we formulate learning as the state-dependent regulation of existing neuromuscular control mechanisms. In this formulation, neuromuscular control is responsible for coordinating muscle activity, while reinforcement learning adjusts these control mechanisms according to the current state, enabling adaptation without relearning muscle coordination from scratch.


Motivated by this insight, we propose a Reflex-Informed Neuromuscular Learning framework for muscle-driven locomotion, where a fixed phase-dependent reflex controller provides the underlying neuromuscular control mechanism. Rather than directly outputting 18-dimensional muscle actions, the reinforcement learning policy produces four biomechanically meaningful residual signals at each control step to modulate the gains and thresholds associated with hip swing, knee support, and ankle propulsion in a state-dependent manner. This provides a compact, interpretable, and biologically meaningful learning interface for muscle-driven locomotion.
To verify that the proposed Reflex-Informed Neuromuscular Learning framework provides both physiological plausibility and adaptability, we conduct complementary experiments under nominal walking, muscle weakness, and external perturbation conditions. Under nominal walking conditions, physiological plausibility is evaluated through human reference kinematics, ground reaction forces, bilateral symmetry, and stride-to-stride consistency. Muscle weakness and external perturbation experiments further evaluate the policy's ability to adapt to changes in musculoskeletal conditions and the environment without retraining. The main contributions of this work are summarized as follows:
\begin{enumerate}

\item We propose a Reflex-Informed Neuromuscular Reinforcement Learning framework that reformulates reinforcement learning as the state-dependent regulation of existing neuromuscular control mechanisms, rather than directly learning muscle actions.
\item We introduce a compact and interpretable learning interface that modulates key pathways of a fixed reflex controller through four biomechanically meaningful residual parameters, enabling neuromuscular regulation without directly controlling individual muscles.
\item We demonstrate that the proposed framework improves physiological plausibility while maintaining adaptability across different locomotion conditions. Extensive experiments show more human-like joint kinematics, ground reaction forces, and gait consistency across multiple walking speeds, while the same trained policy remains effective under muscle weakness and external perturbations without retraining.

\end{enumerate}

\section{Related Work}
\label{sec:related-work}

\subsection{Muscle-Driven Character Simulation}
Musculoskeletal models driven by muscle--tendon dynamics have become a standard framework for studying and generating human locomotion by explicitly modeling muscle activation, muscle--tendon dynamics, skeletal dynamics, and body--environment interaction. In biomechanics, these models have been widely used to generate and analyze healthy and pathological gait, investigate locomotor strategies, study energetic objectives, and evaluate the effects of ageing and musculoskeletal impairments \cite{anderson2001,ackermann2010,falisse2019,song2018,degroote2021,uchida2020,ezati2019,dembia2020}. In computer graphics, muscle-driven simulation has been adopted for physically based character animation, enabling realistic locomotion through muscle-actuated simulation and control \cite{wang2010,wang2012,geijtenbeek2013,lee2019}. Together, these studies establish muscle-driven simulation as a mature foundation for generating realistic human locomotion. Consequently, recent research has increasingly focused on how to design effective controllers for these musculoskeletal systems.

\subsection{Structured Neuromuscular Control}
Biologically inspired controllers introduce prior knowledge of human motor control directly into the control architecture. Representative approaches include central pattern generators (CPGs), muscle synergies, and reflex-based controllers. Ijspeert reviewed CPG-based locomotion control in animals and robots, where rhythmic movement is generated through coupled oscillatory networks \cite{ijspeert2008}. d'Avella et al. and Meyer et al. modeled muscle coordination using muscle synergies, representing muscle activation with a reduced set of coordinated control signals \cite{davella2003,meyer2016}. Geyer and Herr proposed a reflex-based controller that generates human walking through physiologically motivated sensory feedback \cite{geyer2010}, which was later extended by Song and Geyer to generate diverse locomotion behaviors across different walking speeds \cite{song2015}. Dzeladini et al. further combined CPGs and reflexes within a unified neuromuscular controller \cite{dzeladini2014}.

Among these approaches, reflex-based control is particularly relevant to muscle-driven locomotion because it directly maps muscle sensory feedback to muscle stimulation. Reflex-based controllers have been shown to generate stable locomotion in humans \cite{geyer2010,song2015}, adaptive walking across a wide range of speeds \cite{koseki2024}, and robust locomotion in bipedal robots \cite{batts2015}. However, most existing reflex controllers rely on manually designed or offline-optimized parameters. Although muscle stimulation changes continuously with sensory feedback, the underlying reflex gains and thresholds are not updated online, limiting adaptation to changes in musculoskeletal conditions and the environment.
\subsection{Learning-Based Muscle Control}
Deep reinforcement learning has enabled physics-based characters to acquire locomotion skills directly through interaction with the environment, while prior work has shown that the action representation and reference-motion objective strongly affect the learned behavior \cite{peng2017,peng2018}. For muscle-driven characters, existing approaches typically formulate muscle activations as the policy output, allowing muscle control signals to be generated online according to the current state \cite{weng2021,devree2021,ogum2024,song2021}. Compared with manually designed controllers, these methods improve adaptability and can learn complex locomotion behaviors without explicitly specifying muscle coordination.

Recent studies have further improved direct muscle-control reinforcement learning from different perspectives. Bio-inspired reward functions encourage more physiologically realistic gait patterns \cite{nowakowski2021,schumacher2025}. Model-based reinforcement learning incorporates musculoskeletal dynamics into policy optimization
\cite{su2023}. DEP-RL improves exploration in overactuated musculoskeletal systems through embodied exploratory feedback \cite{deprl2023}. More recently, AI-CPG combines a CPG-based feedforward controller with a learned reflex network to improve adaptive locomotion \cite{li2024}.

Overall, these studies indicate that existing approaches cannot simultaneously provide physiological plausibility and adaptability. Structured neuromuscular controllers can generate physiologically plausible movement but struggle to adapt to changes in the musculoskeletal model or the environment, whereas reinforcement learning-based muscle control improves adaptation in muscle-driven locomotion but often generates gait patterns that deviate from human kinematics and dynamics.

\section{Method}
\label{sec:method}
\subsection{Framework Overview}
\begin{figure*}[!t]
\centerline{\includegraphics[width=\textwidth]{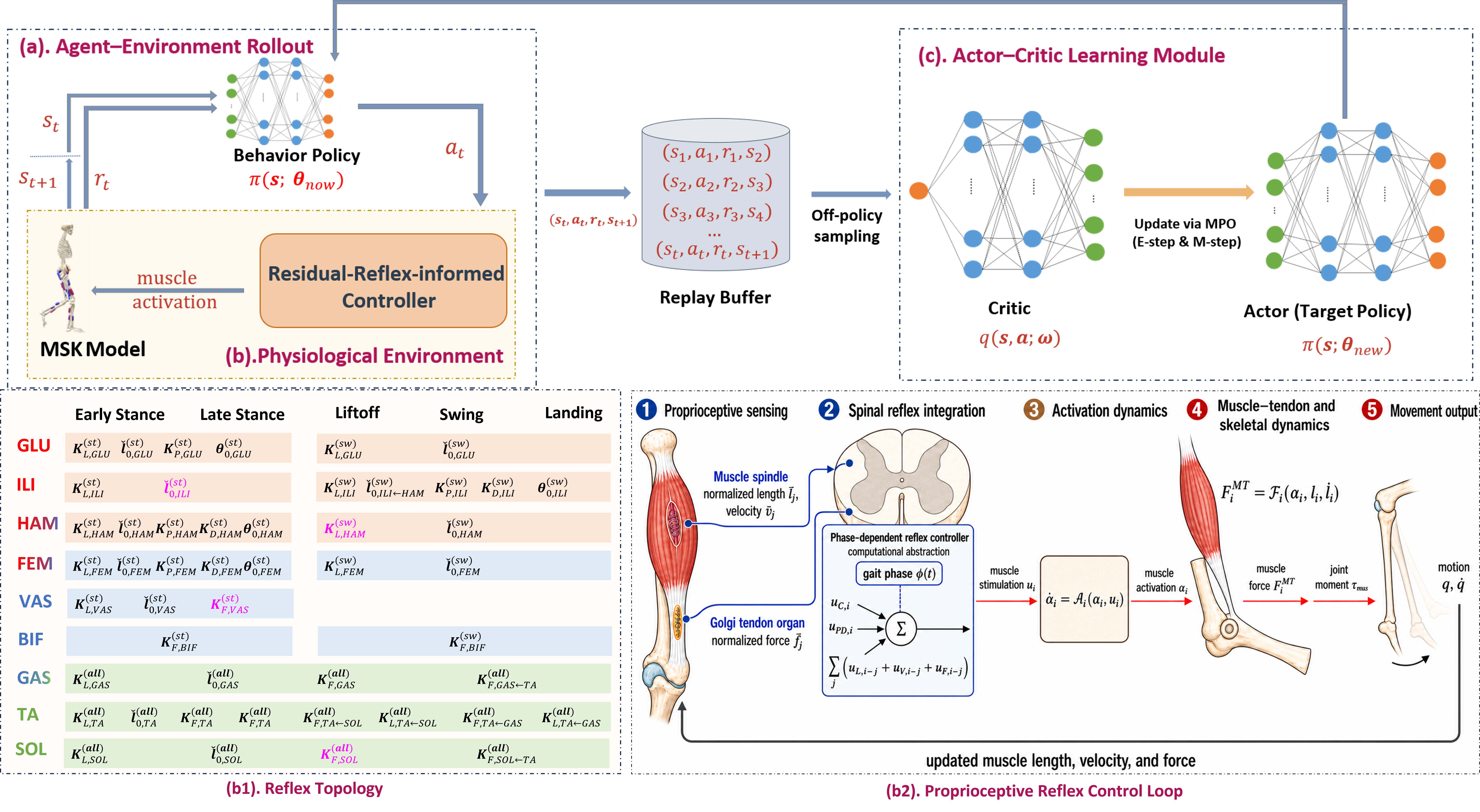}}
\caption{Overview of the proposed Residual-Reflex RL framework. (a) The
rollout policy maps the musculoskeletal state $\mathbf{s}_t$ to a
four-dimensional residual action $\mathbf{a}_t$ and collects transitions
$(\mathbf{s}_t,\mathbf{a}_t,r_t,\mathbf{s}_{t+1})$. (b) The physiological
environment combines the muscle-driven character with a phase-dependent
residual-reflex-informed controller. In the enlarged lower row, panel (b1) on
the left shows the sparse phase-dependent reflex topology, while panel (b2) on
the right shows the proprioceptive control loop from muscle sensing to
stimulation, activation, muscle--tendon force, skeletal motion, and updated
sensory feedback. Magenta entries in (b1) denote the four reflex parameters
modulated by $\mathbf{a}_t$. Superscripts $(\mathrm{all})$, $(\mathrm{st})$,
and $(\mathrm{sw})$ identify parameters shared across all five phases,
Early Stance--Late Stance, and Liftoff--Swing--Landing, respectively. The
notation $i\leftarrow j$ indicates that feedback from source muscle $j$
modulates target muscle $i$; entries without a source are self-feedback
pathways. (c) The
collected transitions are stored in $\mathcal{D}$; the critic
    $q(\mathbf{s},\mathbf{a};\boldsymbol{\omega})$ evaluates candidate actions
    and MPO updates the rollout policy from
    $\pi(\mathbf{s};\boldsymbol{\theta}_{\mathrm{now}})$ to
    $\pi(\mathbf{s};\boldsymbol{\theta}_{\mathrm{new}})$.}
\Description{The framework couples a four-dimensional residual-reflex policy
to a muscle-driven character. The lower row shows the reflex topology on the
left and the proprioceptive reflex control loop on the right. An actor--critic
module learns from replayed transitions.}
\label{framwork}
\end{figure*}

Figure~\ref{framwork} illustrates the proposed Reflex-Informed Neuromuscular Reinforcement Learning framework. A fixed phase-dependent reflex controller provides the underlying neuromuscular control mechanism, while the reinforcement learning policy generates four-dimensional residual actions that regulate selected reflex parameters according to the current musculoskeletal state. The resulting muscle stimulation drives the musculoskeletal simulation, which produces the next state and reward. During learning, transitions $(\mathbf{s}_t,\mathbf{a}_t,r_t,\mathbf{s}_{t+1})$
are stored in the replay buffer $\mathcal{D}$. Maximum a Posteriori Policy Optimization (MPO) updates the policy through an off-policy actor--critic framework using replayed transitions \cite{mpo2018}.



\subsection{Musculoskeletal Model}
\label{sec:musculoskeletal-model}
\begin{figure*}[t]
    \centering
    \includegraphics[width=0.76\textwidth]{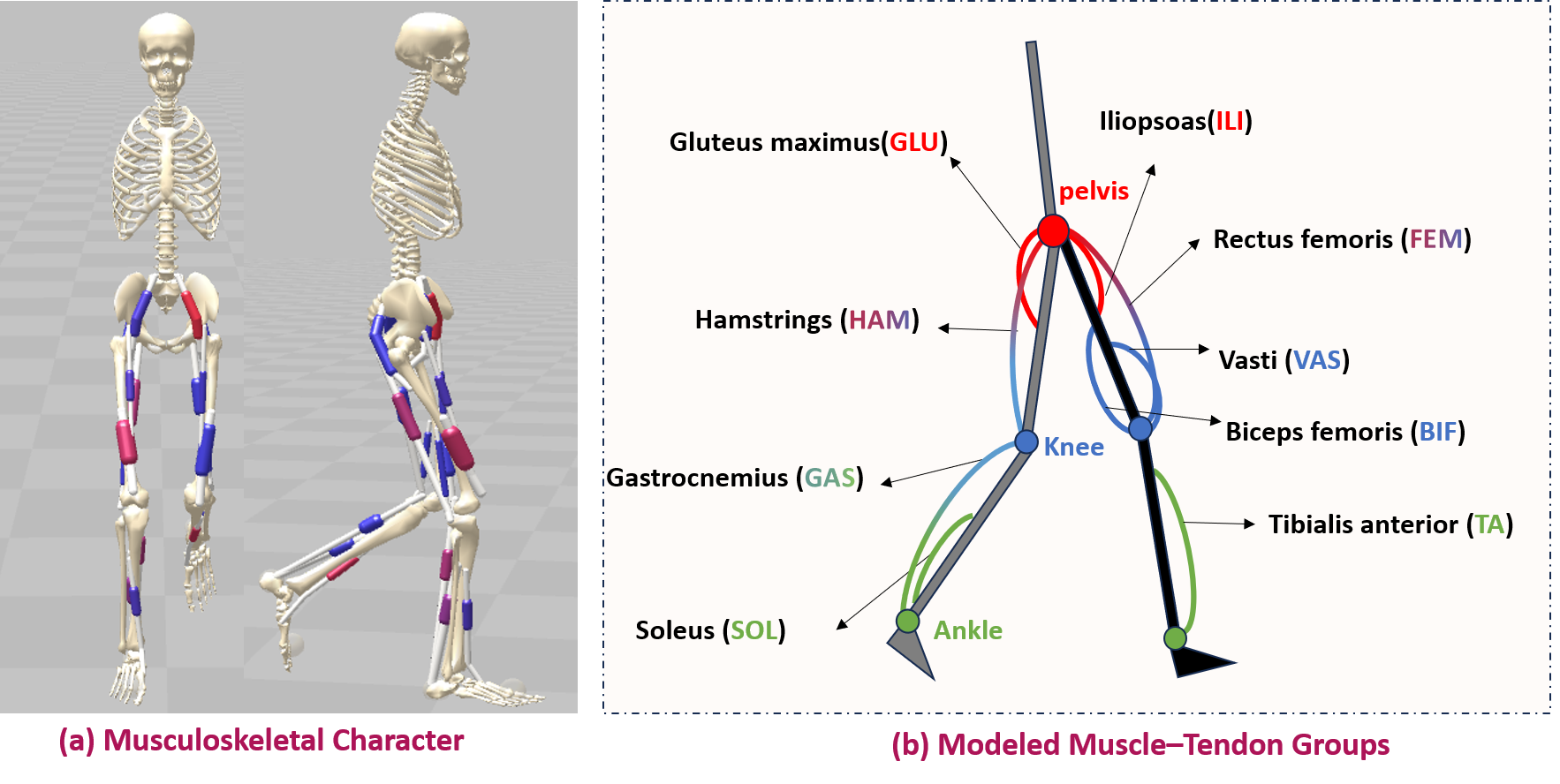}
    \caption{Muscle-driven character model. (a) Front and sagittal views of
    the H0918v2j character used in the experiments. (b) Schematic arrangement
    and primary joint functions of the nine bilateral muscle--tendon groups.
    The complete character is actuated by 18 Hill-type muscle--tendon units,
    with one instance of each modeled group on each leg.}
    \Description{Front and side views of the musculoskeletal character beside
    a schematic of the nine bilateral muscle--tendon groups spanning the hip,
    knee, and ankle.}
    \label{fig:msk-character}
\end{figure*}

\subsubsection{Simulation platform and character.}
The physiological environment module (b) is implemented in the Hyfydy musculoskeletal simulation engine \cite{hyfydy} through the Python-based SCONE Gym interface \cite{geijtenbeek2019}. We use the sagittal-plane H0918v2j human model, which consists of a pelvis, a rigidly attached torso, and bilateral thigh, shank, and foot segments with nine generalized coordinates. The character is actuated by 18 Hill-type muscle--tendon units representing nine muscle groups on each leg, including tibialis anterior (TA), soleus (SOL), gastrocnemius (GAS), vasti (VAS), biceps femoris short head (BIF), rectus femoris (FEM), iliopsoas (ILI), hamstrings (HAM), and gluteus maximus (GLU). Each foot interacts with the ground through spherical heel and toe contact elements, and the controller is updated at 0.005 s intervals. Figure~\ref{fig:msk-character}(a) illustrates the musculoskeletal character, while Fig.~\ref{fig:msk-character}(b) summarizes the arrangement of the modeled muscle--tendon units and the joints they span.


\subsubsection{Muscle-driven dynamics.}
Within module (b), reflex-generated stimulation is applied to the character
through activation dynamics, muscle--tendon mechanics, and forward dynamics. For muscle $i$, neural stimulation $u_i$ is first filtered by
activation dynamics to obtain activation $\alpha_i$, after which the Hill-type muscle model computes the corresponding muscle--tendon force
\begin{equation}
\dot{\alpha}_i=\mathcal{A}_i(\alpha_i,u_i), \qquad
F_i^{\mathrm{MT}}=\mathcal{F}_i(\alpha_i,l_i,\dot l_i),
\end{equation}
where $\mathcal{A}_i$ denotes the activation dynamics and $\mathcal{F}_i$
denotes the Hill-type muscle model. Muscle forces are then transformed into generalized joint moments through configuration-dependent muscle moment arms,
\begin{equation}
\tau_{\mathrm{mus},k}=
\sum_{i\in\mathcal{M}_k}r_{i,k}(\mathbf{q})F_i^{\mathrm{MT}},
\end{equation}
where $r_{i,k}(\mathbf{q})$ is the configuration-dependent muscle moment arm of muscle i about generalized coordinate
$k$, and $\mathcal{M}_k$ denotes the set of muscles spanning that coordinate. Finally, the generalized muscle moments drive the musculoskeletal system through forward dynamics,
\begin{equation}
\mathbf{M}(\mathbf{q})\ddot{\mathbf{q}}+
\mathbf{h}(\mathbf{q},\dot{\mathbf{q}})=
\boldsymbol{\tau}_{\mathrm{mus}}+
\mathbf{J}_{c}^{\mathsf{T}}\mathbf{f}_{c},
\end{equation}
where $\mathbf{q}$, $\dot{\mathbf{q}}$, and $\ddot{\mathbf{q}}$ are the
generalized coordinates, velocities, and accelerations, respectively;
$\mathbf{M}(\mathbf{q})$ is the mass matrix; $\mathbf{h}(\mathbf{q},\dot{\mathbf{q}})$ collects gravitational, Coriolis,
and centrifugal terms; $\boldsymbol{\tau}_{\mathrm{mus}}$ denotes the generalized joint moments generated by the muscles; $\mathbf{J}_{c}$ is the contact Jacobian; and $\mathbf{f}_{c}$ denotes the foot--ground contact forces.

\subsection{Neuromuscular Control Mechanism}
\subsubsection{Physiological feedback}
The reflex controller computes muscle stimulation from delayed muscle sensory feedback and posture feedback. Muscle length and velocity approximate muscle-spindle feedback, whereas muscle force approximates Golgi tendon organ feedback \cite{grillner2020}. These sensory signals form the physiological inputs to the reflex controller. The corresponding gains, thresholds, source muscles, target muscles, and active gait phases are defined in the following sections. As illustrated in Fig.~\ref{framwork}(b2), the resulting muscle stimulation drives activation and muscle--tendon force, while the updated muscle states provide sensory feedback for the next control step.
\subsubsection{Phase-dependent topology.}
A finite-state controller divides the gait cycle into five phases: \textbf{Early Stance}, \textbf{Late Stance}, \textbf{Liftoff}, \textbf{Swing}, and \textbf{Landing}. Reflex pathways are organized into an all-phase group, a stance group, and a swing-related group:
\begin{equation*}
\begin{aligned}
\Phi_{\mathrm{all}} &= \{\mathrm{ES,LS,LF,SW,LA}\},\\
\Phi_{\mathrm{st}}  &= \{\mathrm{ES,LS}\},\\
\Phi_{\mathrm{sw}}  &= \{\mathrm{LF,SW,LA}\}.
\end{aligned}
\end{equation*}
Parameters are shared among phases within the same group but remain independent across groups. For example, the BIF force-feedback gain used during Early Stance and Late Stance differs from that used during Liftoff, Swing, and Landing, allowing the same sensory feedback to support different functions, including stance support, push-off, and swing control, without requiring a separate parameter for every gait phase. Fig.~\ref{framwork}(b1) summarizes the resulting topology across the nine modeled muscle groups. Monoarticular pathways involve TA and SOL at the ankle, VAS and BIF at the knee, and ILI and GLU at the hip. The biarticular muscles GAS, FEM, and HAM span two adjacent joints and therefore contribute to the control of both.

The active parameter-sharing groups are determined by the current gait phase. During Early Stance and Late Stance, the controller activates the all-phase and stance-specific pathways, whereas during Liftoff, Swing, and Landing, it activates the all-phase and swing-related pathways. For convenience, let $\phi(t)$ denote the current gait phase and $\Gamma(\phi)$ the corresponding set of active parameter-sharing groups, such that $\Gamma(\phi)=\{\mathrm{all},\mathrm{st}\}$ during Early Stance and $\Gamma(\phi)=\{\mathrm{all},\mathrm{sw}\}$ during Liftoff, Swing, and Landing.

Each active parameter-sharing group $g\in\Gamma(\phi)$ defines a sparse directed feedback graph $\mathcal{G}_{g}$. A connection $i\leftarrow j$ uses sensory feedback measured from source muscle $j$ to modulate the stimulation of target muscle $i$. Connections with $i=j$ implement self-feedback, whereas connections with $i\neq j$  encode inter-muscle coordination. Reflex gains and thresholds are indexed by the active parameter-sharing group, while inactive pathways have zero gain during the corresponding gait phases. The sign and magnitude of each gain determine how strongly the corresponding sensory feedback facilitates or suppresses the target muscle stimulation.

\subsubsection{Reflex control laws.}
For an active group $g\in\Gamma(\phi(t))$, muscle stimulation is computed from muscle length, velocity, and force feedback.
\begin{align}
u_{L,i\leftarrow j}^{(g)}(t) &= K_{L,i\leftarrow j}^{(g)}
\max(0,\tilde{l}_j(t-t_D)-\tilde{l}_{0,i\leftarrow j}^{(g)}),\\
u_{V,i\leftarrow j}^{(g)}(t) &= K_{V,i\leftarrow j}^{(g)}
\max(0,\tilde{v}_j(t-t_D)-\tilde{v}_{0,i\leftarrow j}^{(g)}),\\
u_{F,i\leftarrow j}^{(g)}(t) &= K_{F,i\leftarrow j}^{(g)}
(\tilde{f}_j(t-t_D)-\tilde{f}_{0,i\leftarrow j}^{(g)}),
\end{align}
where $\tilde{l}_j$, $\tilde{v}_j$, and $\tilde{f}_j$ denote delayed normalized muscle length, velocity, and force measured from source muscle $j$, respectively. The corresponding gains $K_{L}$, $K_{V}$, and $K_{F}$ and thresholds $\tilde{l}_{0}$, $\tilde{v}_{0}$, and $\tilde{f}_{0}$ are indexed by the active parameter-sharing group $g$, and $t_D$ is the sensorimotor delay.

A reflex-like postural feedback term is also used to couple trunk pitch
errors to selected lower-limb muscles.
\begin{equation}
u_{\text{PD},i}^{(g)}(t)=
K_{P,i}^{(g)}(\theta(t-t_D)-\theta_{0,i}^{(g)})+
K_{D,i}^{(g)}\dot{\theta}(t-t_D).
\end{equation}
where $\theta$ and $\dot{\theta}$ are the delayed trunk pitch angle and angular
velocity, $\theta_{0,i}^{(g)}$ is the target pitch offset for muscle $i$, and
$K_{P,i}^{(g)}$ and $K_{D,i}^{(g)}$ are the proportional and derivative
postural-feedback gains for the active group.

The total muscle stimulation of muscle $i$ is obtained by summing the constant background stimulation, muscle sensory feedback and postural feedback contributions from all active parameter-sharing groups.
\begin{equation}
\begin{aligned}
u_i(t) = \sum_{g\in\Gamma(\phi(t))}\Bigg[
&u_{C,i}^{(g)}+u_{\text{PD},i}^{(g)}(t)\\
&+\sum_{j\in\mathcal{M}_{g}}\left(u_{L,i\leftarrow j}^{(g)}(t)
+u_{V,i\leftarrow j}^{(g)}(t)+u_{F,i\leftarrow j}^{(g)}(t)\right)\Bigg].
\end{aligned}
\end{equation}
where $u_{C,i}^{(g)}$ is the group-dependent constant background stimulation and
$\mathcal{M}_{g}$ is the set of source muscles connected to target $i$ in the
corresponding active group. The resulting muscle stimulation drives the activation dynamics and musculoskeletal model described in Sec.~\ref{sec:musculoskeletal-model}.

\subsection{Residual Neuromuscular Regulation}

\subsubsection{Residual formulation}
Rather than directly generating muscle stimulation, the reinforcement learning policy regulates the underlying neuromuscular controller through a four-dimensional residual action $\mathbf{a}_t\in\mathbb{R}^{4}$. Each action component modulates one selected reflex parameter around its nominal value,
\begin{equation}
p_i(t)=p_{i,0}(1+\lambda a_i(t)),
\label{eq:residual_scaling}
\end{equation}
where $p_{i,0}$ is the nominal reflex parameter and $\lambda$  is the action scale. In the present implementation, $\lambda=0.5$. When $a_i(t)=0$, the nominal reflex controller is recovered exactly, whereas nonzero actions produce state-dependent modulation of the selected reflex parameters. The resulting muscle stimulation is then computed by the underlying neuromuscular controller using the current muscle sensory feedback, gait phase, and the regulated reflex parameters.

\begin{table}[!t]
\caption{Selected Residual-Reflex Actions}
\label{tab:residual_actions}
\centering
\small
\setlength{\tabcolsep}{3pt}
\renewcommand{\arraystretch}{1.12}
\begin{tabularx}{\columnwidth}{c c c X}
\toprule
\textbf{Action} & \textbf{Parameter} & \textbf{Pathway} & \textbf{Functional role} \\
\midrule
$a_1$ & $K_{L,\mathrm{HAM}}^{(\mathrm{sw})}$ & S11100.hamstrings.KL & Hip extension / knee flexion \\
$a_2$ & $\tilde{l}_{0,\mathrm{ILI}}^{(\mathrm{st})}$ & S00011.iliopsoas.L0 & Hip flexion / leg advancement \\
\textbf{$a_3$} & \textbf{$K_{F,\mathrm{SOL}}^{(\mathrm{all})}$} & \textbf{S11111.soleus.KF} & \textbf{Ankle support / propulsion} \\
$a_4$ & $K_{F,\mathrm{VAS}}^{(\mathrm{st})}$ & S00011.vasti.KF & Knee extension / stance support \\
\bottomrule
\end{tabularx}
\end{table}

\subsubsection{Regulated reflex parameters}

The four regulated parameters are highlighted in bright magenta in Fig.~\ref{framwork}(b1), and their corresponding pathways are summarized in Table~\ref{tab:residual_actions}. In Table~\ref{tab:residual_actions}, pathway labels follow the naming convention of the controller implementation, while the superscripts (all), (st), and (sw) explicitly indicate the corresponding phase groups used in this paper. Specifically, $a_1$ modulates the swing-related hamstrings length gain to regulate hip extension and knee flexion during leg swing, $a_2$ adjusts the stance-phase iliopsoas length threshold for hip flexion and leg advancement, $a_3$ modulates the all-phase soleus force-feedback gain for ankle support and propulsion, and $a_4$ regulates the stance-phase vasti force-feedback gain for knee extension and stance support.

These four parameters were selected because they regulate gait-critical functions across the major lower-limb joints while directly influencing the key biomechanical roles required for locomotion, including leg swing, stance support, and propulsion. By restricting policy outputs to these biomechanically meaningful reflex parameters, the proposed formulation concentrates policy exploration on neuromuscular mechanisms that are directly related to locomotor function while maintaining a compact and interpretable action space.

\subsection{Policy Learning with MPO}

Policy learning follows an off-policy actor--critic framework optimized with Maximum a Posteriori Policy Optimization (MPO) \cite{mpo2018}. As illustrated in Fig.~\ref{framwork}(c), the actor receives the current musculoskeletal observation and outputs a four-dimensional residual action, while the critic evaluates the corresponding state--action pair. Let $\mathbf{s}_t\in\mathbb{R}^{121}$  denote the observation at time $t$,
\begin{equation}
\begin{aligned}
\mathbf{s}_t = [
&\mathbf{l}_t,\dot{\mathbf{l}}_t,\mathbf{f}_t,\mathbf{e}_t,\boldsymbol{\alpha}_t,\mathbf{p}_{feet,t},\mathbf{q}_t,\dot{\mathbf{q}}_t].
\end{aligned}
\end{equation}
Here $\mathbf{l}$, $\dot{\mathbf{l}}$, and $\mathbf{f}$ are muscle-fiber
lengths, velocities, and forces; $\mathbf{e}$ and $\boldsymbol{\alpha}$ are
muscle excitations and activations; $\mathbf{p}_{feet}$ represents foot
positions relative to the pelvis; $\mathbf{q}$ and
$\dot{\mathbf{q}}$ denote the pelvis and joint degrees of freedom and their corresponding velocities, respectively.

The actor and critic are implemented as multilayer perceptrons with two hidden layers of 256 ReLU units. The actor $\pi(\mathbf{s};\boldsymbol{\theta}_{\mathrm{now}})$ predicts the mean and standard deviation of a Gaussian policy, from which a four-dimensional residual action is sampled during training. During evaluation, the deterministic policy uses the action mean. The critic $q(\mathbf{s},\mathbf{a};\boldsymbol{\omega})$ estimates the corresponding action-value function $Q(\mathbf{s},\mathbf{a})$.

Transitions $(\mathbf{s}_t,\mathbf{a}_t,r_t,\mathbf{s}_{t+1})$ are stored in a replay buffer $\mathcal{D}$. MPO updates the critic from replayed transitions and optimizes the actor under a KL-divergence constraint, enabling stable off-policy learning while reusing collected experience. We follow the standard MPO update procedure and therefore omit the derivation here.

Policy optimization is driven by the following reward function, which consists of a target-speed reward $r_v$ together with penalty terms for excessive contact loading ($c_{grf}$), temporal excitation variation ($c_e$), unnecessary muscle recruitment ($c_m$), joint-limit loading ($c_q$), pelvis instability ($c_{\theta},c_{\dot\theta},c_{com}$), and rapid changes in the residual parameter increments ($c_{\Delta p}$)

\begin{equation}
\begin{aligned}
r_t={}&w_v r_v
-\lambda_{grf}c_{grf}-\lambda_e c_e-\lambda_m c_m
-\lambda_q c_q\\
&-\lambda_{\theta}c_{\theta}
-\lambda_{\dot\theta}c_{\dot\theta}
-\lambda_{com}c_{com}-\lambda_{\Delta p}c_{\Delta p},
\end{aligned}
\end{equation}
The target-speed reward is defined as
\begin{equation}
r_v=\begin{cases}
\exp[-(v_x-v^*)^2], & v_x<v^*,\\
1, & v_x\ge v^*.
\end{cases}
\end{equation}
where $v_x$ and $v^*$ denote the forward center-of-mass speed and the commanded speed, respectively. The individual penalty terms are defined as
\begin{equation}
\begin{aligned}
c_{grf}&=[L_c-1.2]_+,\\
c_e&=\frac{1}{N}\|\mathbf e_t-\mathbf e_{t-1}\|_2^2,\\
c_m&=\frac{1}{d_a}\sum_{i=1}^{N}
\mathbb{I}\!\left[\alpha_i>0.15\right],\\
c_q&=\frac{1}{J}\sum_{j=1}^{J}
\operatorname{mean}\!\left(\left|\boldsymbol{\tau}^{\mathrm{lim}}_j\right|\right).
\end{aligned}
\label{eq:reward_cost1}
\end{equation}
where $L_c$ is the total contact load normalized by body weight; $N$ is the
number of muscles; $d_a$ is the action-space dimension; $J$ is the number of
joints; $\boldsymbol{\tau}^{\mathrm{lim}}_j$ is the joint-limit torque of joint
$j$; $\mathbb{I}[\cdot]$ is the indicator function; and
$[x]_+=\max(0,x)$.
\begin{equation}
\begin{aligned}
c_\theta &= (\theta_t-\theta^*)^2,\\
c_{\dot\theta} &= \dot{\theta}_t^{\,2},\\
c_{com} &= (v_{com,y})^2,\\
\boldsymbol{\delta p}_t &= \lambda\mathbf{p}_0\odot\mathbf{a}_t,\\
c_{\Delta p} &= \frac14\|\boldsymbol{\delta p}_t-\boldsymbol{\delta p}_{t-1}\|_2^2.
\end{aligned}
\label{eq:reward_cost3}
\end{equation}
where $\theta^*$ is the target pelvis pitch; $v_{com,y}$ is the vertical center-of-mass velocity; $\mathbf a_t$ denotes the four-dimensional residual action produced by the policy; $\mathbf p_0$ collects the corresponding nominal reflex parameters; $\odot$ denotes element-wise multiplication; and $\boldsymbol{\delta p}_t$ is the vector of residual parameter increments applied at time $t$.

\section{Evaluation}
\label{sec:evaluation-protocol}

\subsection{Evaluation Protocol}
\subsubsection{Experimental Setup}
The simulator, musculoskeletal character, muscle-driven dynamics, and
foot--ground contact model are described in Sec.~\ref{sec:musculoskeletal-model}. The same character and
simulation settings were used for all controllers unless a plantarflexor
weakness condition was explicitly introduced.

The evaluation includes four control approaches: two direct muscle-control
reinforcement learning methods, E2E-RL \cite{weng2021,devree2021,ogum2024}
and DEP-RL \cite{deprl2023}; the proposed Residual-Reflex RL method; and a
reflex controller whose parameters were optimized offline using CMA-ES.
Table~\ref{tab:policy_interface} summarizes the policy interfaces of the
three reinforcement learning methods. The CMA-ES baseline has no learned
policy output and keeps its optimized reflex parameters fixed during
evaluation. All reinforcement learning policies were trained using the same MPO framework and musculoskeletal model. Training was performed on a workstation equipped with an AMD Ryzen Threadripper PRO 9965WX CPU and 256 GB RAM.

\begin{table}[!t]
\centering
\caption{Policy interfaces of the reinforcement learning methods.}
\label{tab:policy_interface}
\small
\setlength{\tabcolsep}{4pt}
\renewcommand{\arraystretch}{1.15}
\begin{tabularx}{\linewidth}{l c X}
\toprule
\textbf{Method} & \textbf{Action dim.} & \textbf{Policy output} \\
\midrule
E2E-RL & 18 & Direct muscle-level control actions. \\
DEP-RL & 18 & Direct muscle-level control with DEP-style embodied exploration. \\
\textbf{Residual-Reflex RL} & \textbf{4} & \textbf{Residual modulation of selected reflex parameters.} \\
\bottomrule
\end{tabularx}
\end{table}

\subsubsection{Tasks and Evaluation Metrics}
The evaluation consists of three complementary experiments designed to assess both physiological plausibility and adaptability. Nominal walking evaluates gait quality under controlled target-speed tasks, plantarflexor weakness evaluates adaptation to changes in musculoskeletal capacity, and external push perturbation evaluates recovery from environmental disturbances.

For nominal walking, policies were evaluated at target walking speeds of 0.8, 1.0, and 1.2 m/s. Each policy was simulated for 25 s, and the resulting trajectories were exported as STO files for post-hoc gait analysis. All detected gait cycles were normalized to 0--100\% of the gait cycle. The comparison includes E2E-RL, DEP-RL, and Residual-Reflex RL. Four groups of metrics are used to evaluate task performance, physiological plausibility, and gait consistency.

\paragraph{Speed tracking and stride measures}
To evaluate task performance, we report
realized walking speed, absolute speed error
$|\hat{v}-v_{\mathrm{target}}|$, stride length, stride time, and the number of analyzed cycles. The realized walking speed and speed error quantify how accurately the controller follows the commanded locomotion task, whereas stride length and stride time characterize the gait strategy adopted to achieve the target speed.

\paragraph{Human-reference kinematic gait score}
To evaluate kinematic plausibility, we compare the generated gait-cycle profiles with human-reference kinematic data for the hip, knee, and ankle joints. These reference profiles are used only for offline evaluation and are not included in the reinforcement learning reward. For each joint, left and right gait cycles were time-normalized and averaged to obtain a mean simulated profile $\bar{y}(x)$, where $x$ denotes the normalized gait-cycle percentage. Let $y_{\min}(x_k)$ and $y_{\max}(x_k)$ denote the lower and upper human-reference bounds at sample point $x_k$. The normalized reference-range violation is
\begin{equation}
e = \frac{1}{N}\sum_{k=1}^{N}
\frac{
\left[y_{\min}(x_k)-\bar{y}(x_k)\right]_+
+ \left[\bar{y}(x_k)-y_{\max}(x_k)\right]_+
}{
\max\left(0.01, y_{\max}(x_k)-y_{\min}(x_k)\right)
},
\label{eq:reference_range_violation}
\end{equation}
where $[z]_+=\max(z,0)$.
The corresponding score is
\begin{equation}
S = 100 \cdot \mathrm{clip}(1-e,0,1).
\end{equation}
Higher scores indicate closer agreement with the human-reference gait profiles. A score of 100 indicates that the averaged trajectory remains entirely within the reference range. The overall kinematic score is computed as the average of the hip, knee, and ankle scores.

\paragraph{Vertical GRF human-reference score}
To evaluate dynamic plausibility, we analyze the vertical ground reaction force (GRF), which directly reflects body loading and push-off behavior. Individual left and right GRF gait cycles are visualized to assess stance-phase loading patterns, double-support and propulsion behavior, and stride-to-stride consistency. The numerical GRF score is computed using the human-reference range metric in Eq.~\ref{eq:reference_range_violation}, applied to the vertical GRF profile, whereas the waveform plots provide a qualitative comparison of left--right symmetry and stride-to-stride repeatability.

\paragraph{Symmetry and repeatability score}
To evaluate gait consistency and repeatability, we quantify bilateral symmetry and stride-to-stride repeatability. Bilateral symmetry is measured as the root-mean-square (RMS) difference between the average left and right normalized gait-cycle waveforms. For a gait-cycle quantity $y$, this waveform difference is
\begin{equation}
D_y =
\sqrt{\frac{1}{N}\sum_{k=1}^{N}
\left(\bar{y}_{L}(x_k)-\bar{y}_{R}(x_k)\right)^2},
\end{equation}
where $y$ denotes a normalized gait-cycle waveform, $x_k$ is the $k$th
gait-cycle sample, $N$ is the number of samples in the normalized gait cycle,
and $\bar{y}_{L}$ and $\bar{y}_{R}$ are the average left- and right-side
profiles computed over the analyzed cycles. In the results, lower $D_y$
indicates more symmetric left--right motion.
Stride-to-stride repeatability is quantified using the coefficient of variation (CV) of stride time and stride length,
\begin{equation}
\mathrm{CV}_{z}=100\frac{\sigma_z}{\mu_z},
\end{equation}
where $z$ denotes either stride time or stride length, $\mu_z$ is the mean
value of $z$, and $\sigma_z$ is its standard deviation over the analyzed gait
cycles. The factor 100 expresses the coefficient of variation as a percentage. Lower $\mathrm{CV}_z$ indicates more repeatable gait cycles.
In the reported results, the Hip, Knee, Ankle, and GRF columns correspond to $D_y$ for the respective left--right waveform pairs, whereas the Time CV and Length CV columns report $\mathrm{CV}_z$ for stride time and stride length, respectively.

\subsection{Nominal Gait Generation}
This experiment evaluates whether the proposed controller generates physiologically plausible locomotion under nominal walking conditions. Three controllers (E2E-RL, DEP-RL, and Residual-Reflex RL) are compared at target walking speeds of 0.8, 1.0, and 1.2 m/s. The evaluation proceeds from task performance to kinematic plausibility, dynamic plausibility, gait consistency, and qualitative motion analysis.

\label{sec:nominal-gait}
\subsubsection{Task Performance}
The speed tracking and stride measures show that all three controllers successfully generated sustained forward walking under the commanded speed tasks. As summarized in Table~\ref{tab:spatiotemporal_metrics}, Residual-Reflex RL achieved the most accurate speed tracking at all target speeds, with absolute speed errors of 0.01, 0.05, and 0.01 m/s at 0.8, 1.0, and 1.2 m/s, respectively. E2E-RL and DEP-RL also completed all walking tasks, but exhibited larger speed errors, particularly at higher target speeds where both methods tended to overshoot the commanded velocity.

These results establish that the subsequent comparisons are performed among controllers that all accomplish the locomotion task, rather than between successful and failed walking trials. The stride measures further reveal different gait strategies. E2E-RL tends to use shorter and faster steps at lower walking speeds, whereas DEP-RL and Residual-Reflex RL adopt longer strides. The following analyses therefore investigate whether successful task execution is also accompanied by human-like kinematics, dynamic consistency, and gait consistency.

\begin{table}[!t]
\centering
\caption{Speed tracking and stride measures from the analyzed walking
trajectories. Speed error is the absolute difference between realized and target
speed. Here $v^*$ and $\hat v$ are the target and realized speeds, $L$ and $T$
are stride length and stride time, and $N$ is the number of analyzed cycles.}
\label{tab:spatiotemporal_metrics}
\normalsize
\setlength{\tabcolsep}{1pt}
\renewcommand{\arraystretch}{1.08}
\begin{tabular*}{\columnwidth}{@{\extracolsep{\fill}}c l r r r r r@{}}
\toprule
\boldmath{$v^*$} & \textbf{Method} & \boldmath{$\hat v$} & \boldmath{$|\Delta v|$} & \boldmath{$L$} & \boldmath{$T$} & \boldmath{$N$} \\
\midrule
\multirow[c]{3}{*}{0.8} & E2E-RL & 0.87 & 0.07 & 0.78 & 0.89 & 50 \\
 & DEP-RL & 0.83 & 0.03 & 1.09 & 1.31 & 33 \\
 & \textbf{Residual-Reflex RL} & 0.79 & \textbf{0.01} & 0.99 & 1.26 & 35 \\
\midrule
\multirow[c]{3}{*}{1.0} & E2E-RL & 1.18 & 0.18 & 1.39 & 1.18 & 37 \\
 & DEP-RL & 1.14 & 0.14 & 1.29 & 1.13 & 39 \\
 & \textbf{Residual-Reflex RL} & 1.05 & \textbf{0.05} & 1.23 & 1.17 & 37 \\
\midrule
\multirow[c]{3}{*}{1.2} & E2E-RL & 1.26 & 0.06 & 1.23 & 0.97 & 46 \\
 & DEP-RL & 1.30 & 0.10 & 1.37 & 1.06 & 42 \\
 & \textbf{Residual-Reflex RL} & 1.21 & \textbf{0.01} & 1.34 & 1.11 & 40 \\
\bottomrule
\end{tabular*}
\end{table}

\subsubsection{Joint Kinematic Plausibility}
\begin{figure}[!t]
\centerline{\includegraphics[width=\columnwidth]{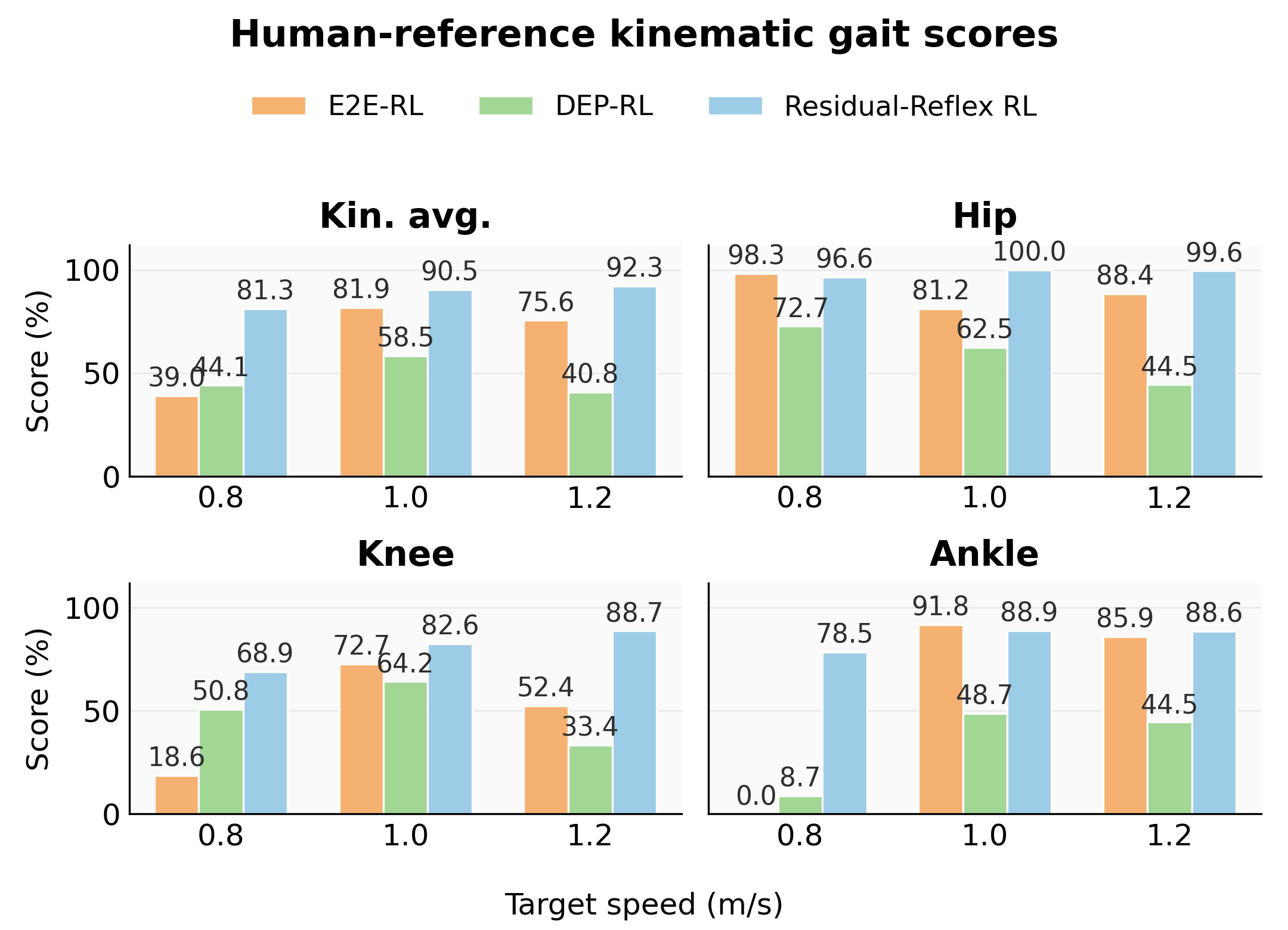}}
\caption{Human-reference kinematic gait scores across target speeds. Numeric
labels indicate the score of each bar. The Residual-Reflex RL controller achieves
the highest kinematic-average score and especially improves knee and ankle
components.}
\Description{Grouped bars compare average hip, knee, and ankle kinematic scores
for three controllers at three walking speeds. Residual-Reflex RL has the
highest kinematic-average score at every speed.}
\label{fig:gait_score_bars}
\end{figure}

The human-reference kinematic scores reveal a clear separation between the three controllers. As summarized in Table~\ref{tab:gait_similarity}, Residual-Reflex RL achieved the highest joint-kinematic score at all target walking speeds, with an average score of 88.0\% across the 0.8, 1.0, and 1.2 m/s evaluations. The improvement is most pronounced in the knee and ankle joints, where the proposed controller consistently preserves human-reference joint coordination. In contrast, E2E-RL maintains relatively high hip scores but exhibits less consistent knee and ankle motion, whereas DEP-RL performs consistently worse, particularly at 1.2 m/s. Fig.~\ref{fig:gait_score_bars} summarizes these quantitative comparisons across all target speeds.

Scores close to zero indicate that the averaged joint trajectory remains outside the human-reference range over most of the gait cycle, rather than exhibiting only localized deviations. This distinction is important because controllers can successfully complete the walking task while still producing joint motions that differ substantially from human gait.

\begin{table}[!t]
\centering
\caption{Human-reference kinematic scores in controlled walking validation.
Scores are computed from normalized hip, knee, and ankle gait-cycle profiles.
The kinematic average (Avg.) is the mean of the hip, knee, and ankle scores;
$v^*$ and $\hat v$ are the target and realized speeds.}
\label{tab:gait_similarity}
\normalsize
\setlength{\tabcolsep}{1pt}
\renewcommand{\arraystretch}{1.08}
\begin{tabular*}{\columnwidth}{@{\extracolsep{\fill}}c l r r r r r@{}}
\toprule
\boldmath{$v^*$} & \textbf{Method} & \boldmath{$\hat v$} & \textbf{Avg.} & \textbf{Hip} & \textbf{Knee} & \textbf{Ankle} \\
\midrule
\multirow[c]{3}{*}{0.8} & E2E-RL & 0.87 & 39.0 & \textbf{98.3} & 18.6 & 0.0 \\
 & DEP-RL & 0.83 & 44.1 & 72.7 & 50.8 & 8.7 \\
 & \textbf{Residual-Reflex RL} & 0.79 & \textbf{81.3} & 96.6 & \textbf{68.9} & \textbf{78.5} \\
\midrule
\multirow[c]{3}{*}{1.0} & E2E-RL & 1.18 & 81.9 & 81.2 & 72.7 & \textbf{91.8} \\
 & DEP-RL & 1.14 & 58.5 & 62.5 & 64.2 & 48.7 \\
 & \textbf{Residual-Reflex RL} & 1.05 & \textbf{90.5} & \textbf{100.0} & \textbf{82.6} & 88.9 \\
\midrule
\multirow[c]{3}{*}{1.2} & E2E-RL & 1.26 & 75.6 & 88.4 & 52.4 & 85.9 \\
 & DEP-RL & 1.30 & 40.8 & 44.5 & 33.4 & 44.5 \\
 & \textbf{Residual-Reflex RL} & 1.21 & \textbf{92.3} & \textbf{99.6} & \textbf{88.7} & \textbf{88.6} \\
\bottomrule
\end{tabular*}
\end{table}

\begin{figure}[!t]
\centerline{\includegraphics[width=\columnwidth]{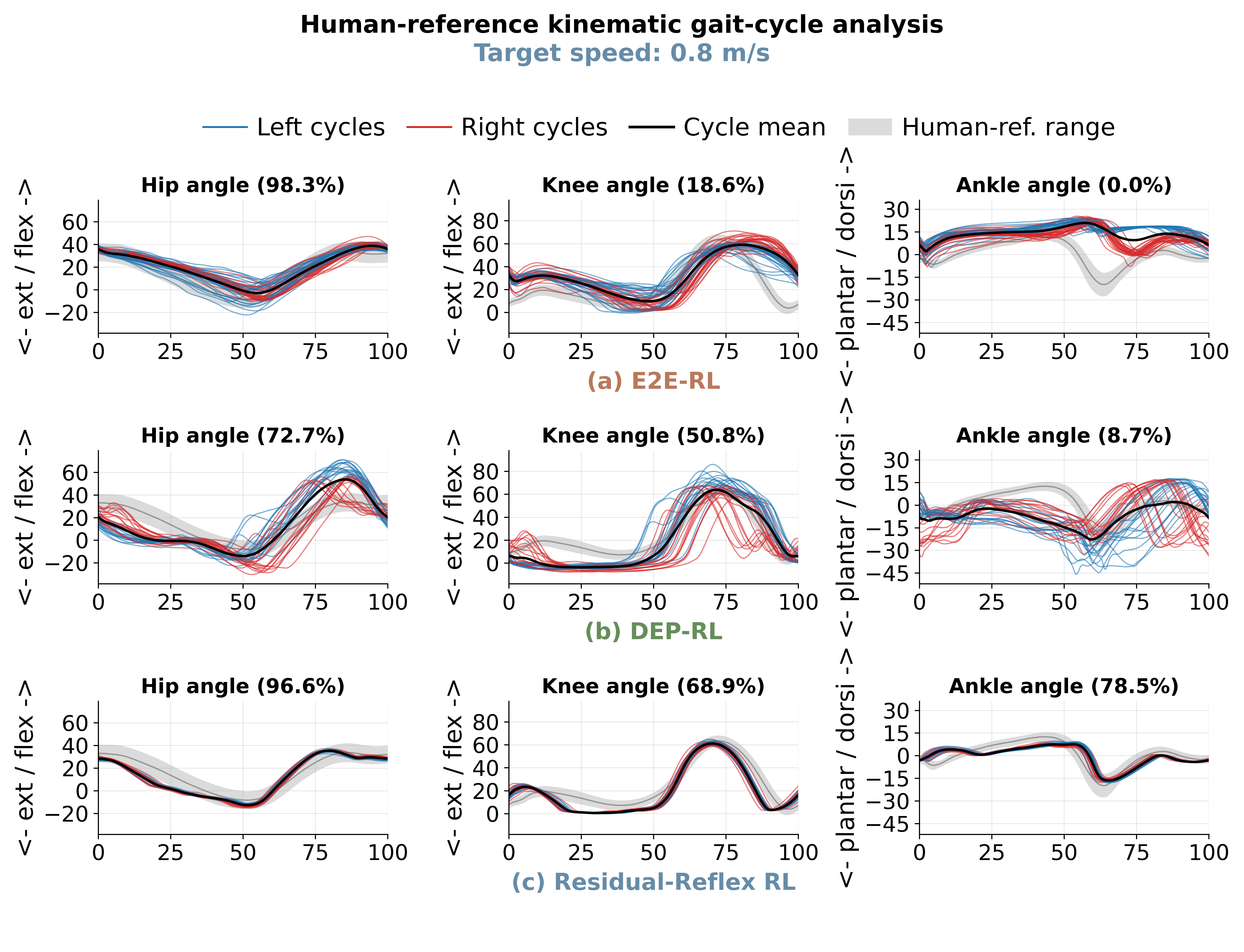}}
\caption{Human-reference kinematic gait-cycle visualization at target speed
\textbf{0.8~m/s} for E2E-RL, DEP-RL, and Residual-Reflex RL. Blue and red
curves denote left and right gait cycles, respectively; the black curve is the
cycle average and the gray band is the human-reference range. Horizontal axes
show normalized gait cycle percentage.}
\Description{Hip, knee, and ankle gait-cycle curves for three controllers at
0.8 meters per second are compared with human-reference ranges.}
\label{fig:gait_profiles_08}
\end{figure}

\begin{figure}[!t]
\centerline{\includegraphics[width=\columnwidth]{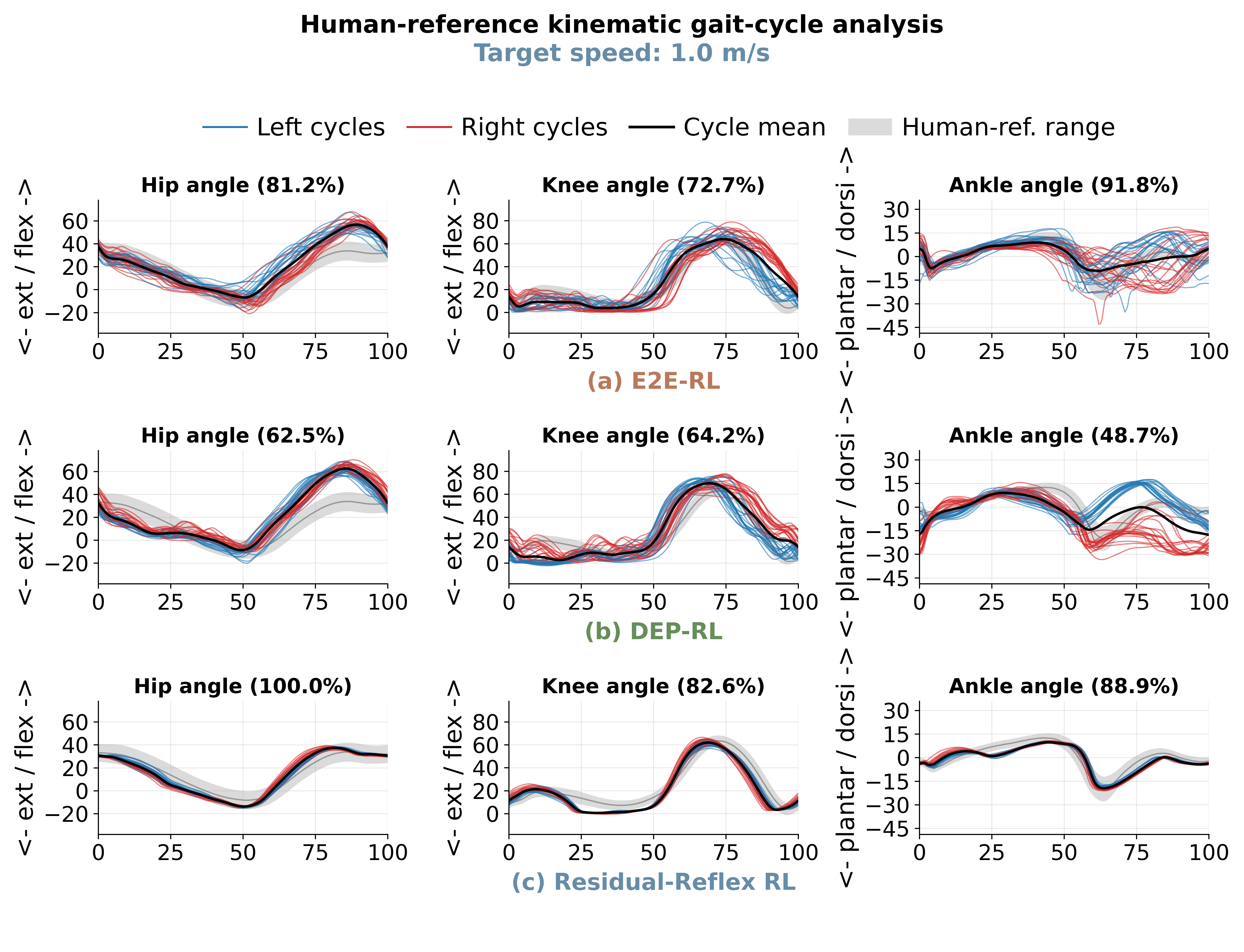}}
\caption{Human-reference kinematic gait-cycle visualization at target speed
\textbf{1.0~m/s} for E2E-RL, DEP-RL, and Residual-Reflex RL. Blue and red
curves denote left and right gait cycles, respectively; the black curve is the
cycle average and the gray band is the human-reference range. Horizontal axes
show normalized gait cycle percentage.}
\Description{Hip, knee, and ankle gait-cycle curves for three controllers at
1.0 meter per second are compared with human-reference ranges.}
\label{fig:gait_profiles_10}
\end{figure}

\begin{figure}[!t]
\centerline{\includegraphics[width=\columnwidth]{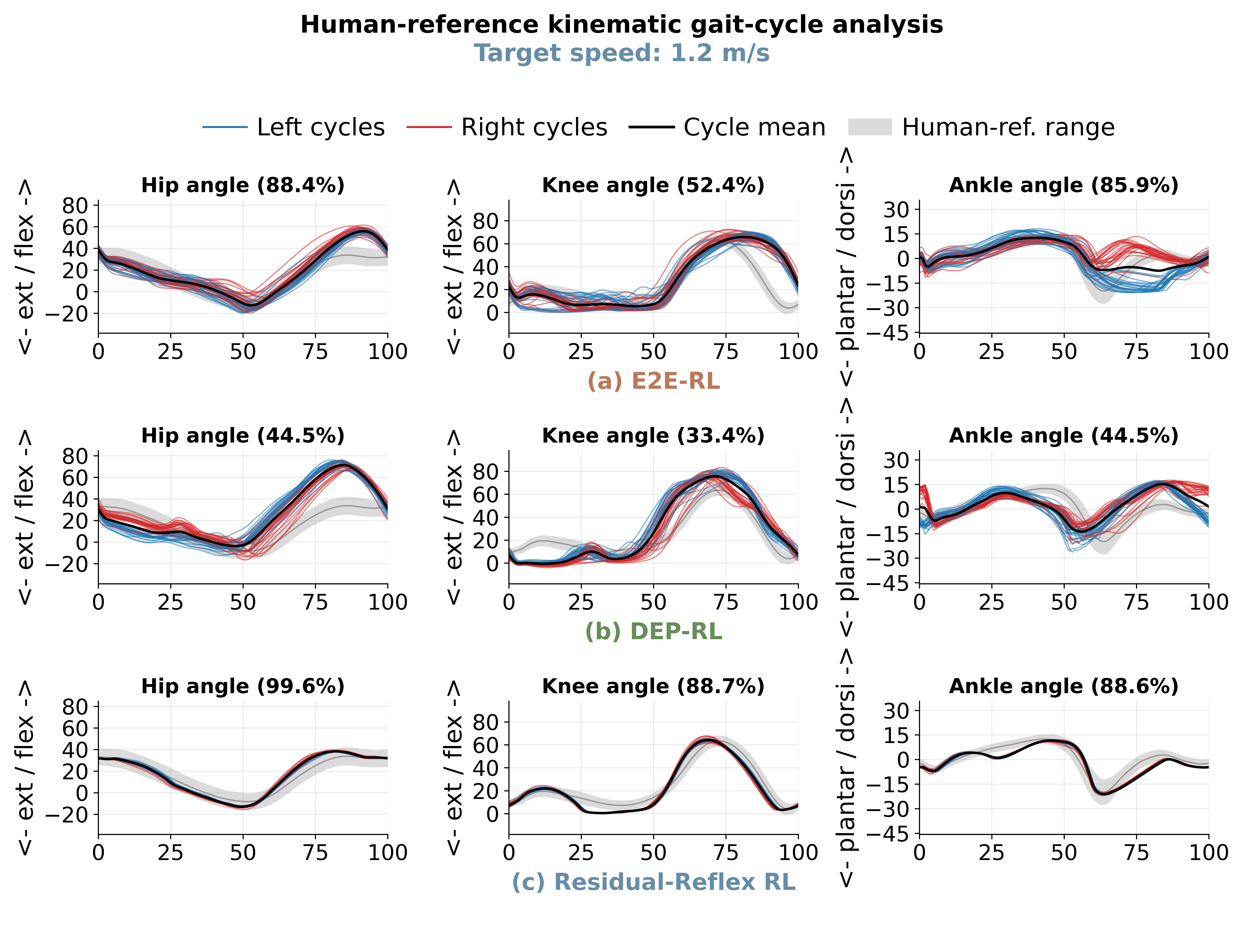}}
\caption{Human-reference kinematic gait-cycle visualization at target speed
\textbf{1.2~m/s} for E2E-RL, DEP-RL, and Residual-Reflex RL. Blue and red
curves denote left and right gait cycles, respectively; the black curve is the
cycle average and the gray band is the human-reference range. Horizontal axes
show normalized gait cycle percentage.}
\Description{Hip, knee, and ankle gait-cycle curves for three controllers at
1.2 meters per second are compared with human-reference ranges.}
\label{fig:gait_profiles_12}
\end{figure}

The waveform comparisons in Figs.~\ref{fig:gait_profiles_08}--\ref{fig:gait_profiles_12} further explain these quantitative results and illustrate how the controllers behave across different target speeds. At 0.8 m/s, E2E-RL produced a reasonable hip trajectory but failed to maintain human-reference knee and ankle profiles, whereas DEP-RL exhibited larger left--right differences and stride-to-stride variability. At 1.0 m/s, all controllers produced gait patterns closer to the reference trajectories, but Residual-Reflex RL still generated the most compact waveform distribution and the highest overall score. At 1.2 m/s, where faster stance-to-swing transitions are required, the difference became more pronounced. Residual-Reflex RL maintained knee and ankle trajectories clustered around the human-reference ranges, whereas E2E-RL and DEP-RL showed larger deviations and more scattered gait cycles.

These observations are consistent with the quantitative scores reported in Table~\ref{tab:gait_similarity} and further explain why Residual-Reflex RL produces more repeatable locomotion. The narrower distributions of the left and right gait cycles indicate more consistent joint coordination across both successive gait cycles and bilateral limbs.

\subsubsection{Dynamic Plausibility}
The vertical GRF scores further distinguish the three controllers. As summarized in Table~\ref{tab:grf_scores}, Residual-Reflex RL consistently achieved the highest vertical GRF scores across all target walking speeds, indicating the closest agreement with the human-reference loading patterns. In contrast, both E2E-RL and DEP-RL produced substantially lower GRF morphology scores, with DEP-RL showing near-zero agreement at 1.0 and 1.2 m/s.

\begin{figure}[!t]
\centerline{\includegraphics[width=\columnwidth]{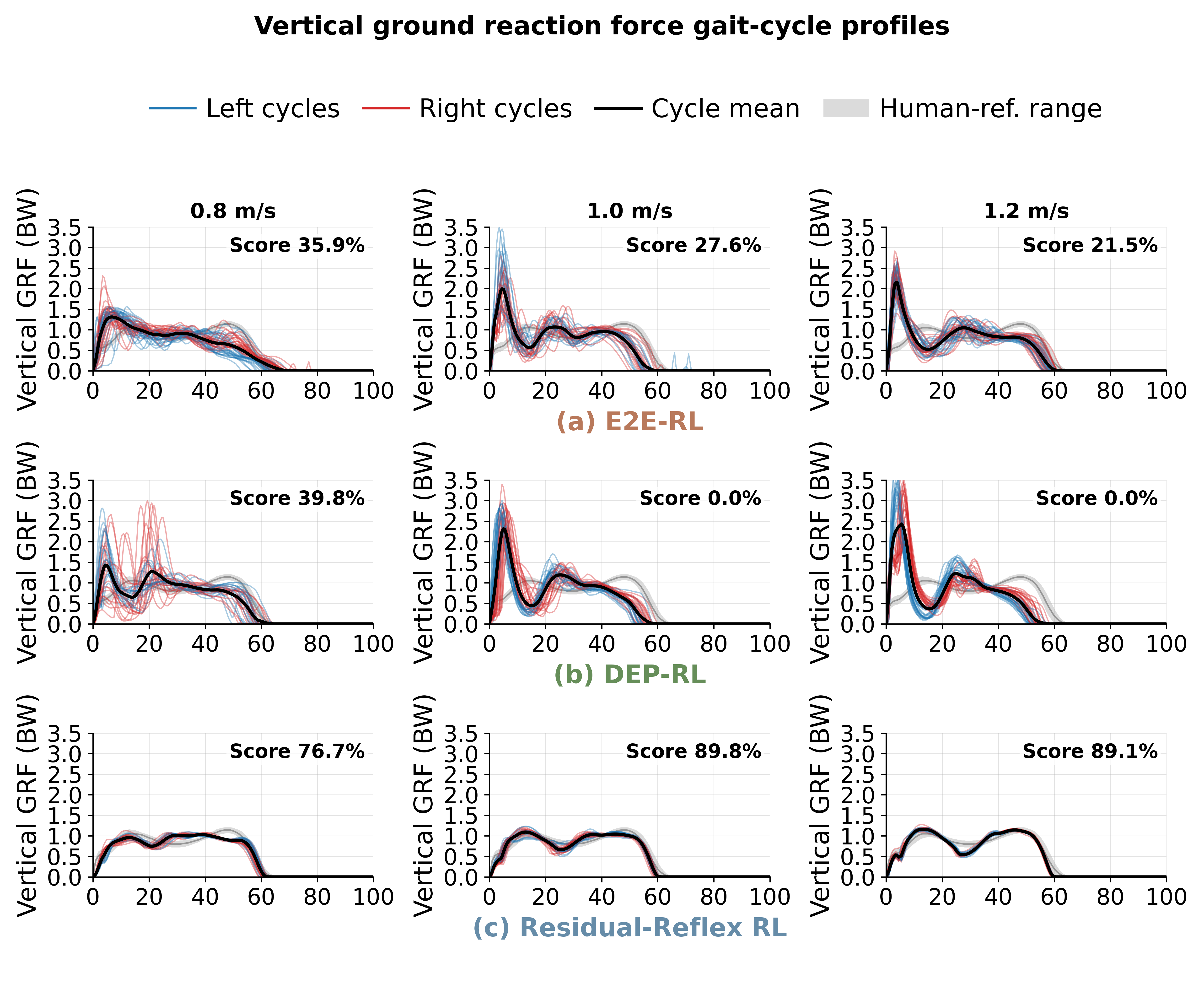}}
\caption{Vertical GRF morphology across target speeds for E2E-RL, DEP-RL, and
Residual-Reflex RL. Blue and red thin curves show individual left
and right normalized gait cycles, respectively; the black curve denotes the
cycle mean, the gray band denotes the human-reference range used for post-hoc
evaluation, and the panel annotations report the GRF human-reference score.
Horizontal axes show normalized gait cycle percentage.}
\Description{A grid compares left and right vertical ground-reaction-force
cycles for three controllers and three target speeds with human-reference
ranges. Residual-Reflex RL produces the most concentrated curves.}
\label{fig:grf_morphology}
\end{figure}

The corresponding GRF waveforms are shown in Fig.~\ref{fig:grf_morphology}. Residual-Reflex RL generated smoother vertical GRF profiles with a more human-reference double-support pattern throughout stance. In contrast, E2E-RL exhibited irregular loading despite producing visually plausible joint trajectories in some cases, whereas DEP-RL produced pronounced early-stance impact peaks and larger stride-to-stride variations, particularly at higher walking speeds.

A GRF score of 0 indicates that the averaged force profile remains outside the human-reference range after normalization, rather than reflecting only localized deviations. This observation highlights an important distinction between kinematic and dynamic evaluation. Although E2E-RL and DEP-RL can generate visually plausible joint motion, their foot--ground interactions remain substantially less consistent with human walking dynamics. The smoother GRF profiles produced by Residual-Reflex RL therefore demonstrate that the proposed controller improves not only joint kinematics but also dynamic consistency during stance.

\begin{table}[!t]
\centering
\caption{Vertical GRF human-reference scores in controlled walking validation.
Higher values indicate closer agreement with the human-reference vertical GRF
range.}
\label{tab:grf_scores}
\small
\setlength{\tabcolsep}{4pt}
\renewcommand{\arraystretch}{1.12}
\begin{tabular*}{\columnwidth}{@{\extracolsep{\fill}}l r r r}
\toprule
\textbf{Method} & \textbf{0.8~m/s} & \textbf{1.0~m/s} & \textbf{1.2~m/s} \\
\midrule
E2E-RL & 35.9 & 27.6 & 21.5 \\
DEP-RL & 39.8 & 0.0 & 0.0 \\
\textbf{Residual-Reflex RL} & \textbf{76.7} & \textbf{89.8} & \textbf{89.1} \\
\bottomrule
\end{tabular*}
\end{table}

\subsubsection{Gait Consistency}
The symmetry and repeatability scores demonstrate that Residual-Reflex RL consistently produces more symmetric and repeatable gait cycles than E2E-RL and DEP-RL. As summarized in Table~\ref{tab:symmetry_variability}, Residual-Reflex RL achieved lower left--right waveform differences and lower stride time and stride length coefficients of variation across all target walking speeds. The Hip, Knee, Ankle, and GRF columns report the left--right waveform difference $D_y$, whereas the Time CV and Length CV columns report the coefficient of variation of stride time and stride length, respectively. The largest improvements are observed in ankle symmetry and stride variability, where E2E-RL and DEP-RL exhibit substantially less repeatable gait cycles.

These quantitative results are consistent with the waveform visualizations, indicating that residual-reflex modulation improves bilateral coordination as well as stride-to-stride repeatability. For example, at 1.2 m/s, Residual-Reflex RL achieved stride time and stride length coefficients of variation of only 0.70\% and 0.74\%, respectively, indicating that the standard deviations of these quantities were below 1\% of their corresponding mean values. In contrast, E2E-RL and DEP-RL exhibited considerably larger cycle-to-cycle variability.

These results indicate that the proposed controller not only produces a physiologically plausible average gait cycle, but also maintains consistent locomotion over successive steps.

\begin{table}[!t]
\centering
\caption{Cycle-level left--right waveform differences and stride variability.
Panel (a) reports $D_y$, the RMS difference between the average left and right
gait-cycle waveforms. Panel (b) reports $100\sigma_z/\mu_z$ for stride time and
stride length. Lower values indicate more symmetric or more repeatable gait.}
\label{tab:symmetry_variability}
\normalsize
\setlength{\tabcolsep}{1pt}
\renewcommand{\arraystretch}{1.05}
\textbf{(a) Left--right waveform differences $D_y$}\par\smallskip
\begin{tabular*}{\columnwidth}{@{\extracolsep{\fill}}c l r r r r@{}}
\toprule
\boldmath{$v^*$} & \textbf{Method} & \textbf{Hip} & \textbf{Knee} & \textbf{Ankle} & \textbf{GRF} \\
\midrule
\multirow[c]{3}{*}{0.8} & E2E-RL & 2.92 & 5.75 & 6.36 & 0.10 \\
 & DEP-RL & 11.20 & 11.73 & 11.59 & 0.12 \\
 & \textbf{Residual-Reflex RL} & \textbf{0.80} & \textbf{1.03} & \textbf{0.89} & \textbf{0.04} \\
\midrule
\multirow[c]{3}{*}{1.0} & E2E-RL & 6.12 & 10.61 & 6.69 & 0.07 \\
 & DEP-RL & 4.54 & 10.12 & 16.27 & 0.28 \\
 & \textbf{Residual-Reflex RL} & \textbf{1.78} & \textbf{1.85} & \textbf{0.97} & \textbf{0.03} \\
\midrule
\multirow[c]{3}{*}{1.2} & E2E-RL & 3.43 & 3.00 & 9.84 & 0.04 \\
 & DEP-RL & 8.33 & 5.83 & 6.31 & 0.34 \\
 & \textbf{Residual-Reflex RL} & \textbf{0.87} & \textbf{0.87} & \textbf{0.46} & \textbf{0.02} \\
\bottomrule
\end{tabular*}

\vspace{4pt}
\textbf{(b) Stride variability (\%)}\par\smallskip
\begin{tabular*}{\columnwidth}{@{\extracolsep{\fill}}c l r r@{}}
\toprule
\boldmath{$v^*$} & \textbf{Method} & \textbf{Time CV} & \textbf{Length CV} \\
\midrule
\multirow[c]{3}{*}{0.8} & E2E-RL & 9.05 & 14.88 \\
 & DEP-RL & 7.96 & 11.04 \\
 & \textbf{Residual-Reflex RL} & \textbf{1.98} & \textbf{5.66} \\
\midrule
\multirow[c]{3}{*}{1.0} & E2E-RL & 5.03 & 5.63 \\
 & DEP-RL & 2.62 & 5.27 \\
 & \textbf{Residual-Reflex RL} & \textbf{1.51} & \textbf{1.86} \\
\midrule
\multirow[c]{3}{*}{1.2} & E2E-RL & 3.87 & 5.83 \\
 & DEP-RL & 2.29 & 4.78 \\
 & \textbf{Residual-Reflex RL} & \textbf{0.70} & \textbf{0.74} \\
\bottomrule
\end{tabular*}
\end{table}

\begin{figure}[!t]
\centerline{\includegraphics[width=\columnwidth]{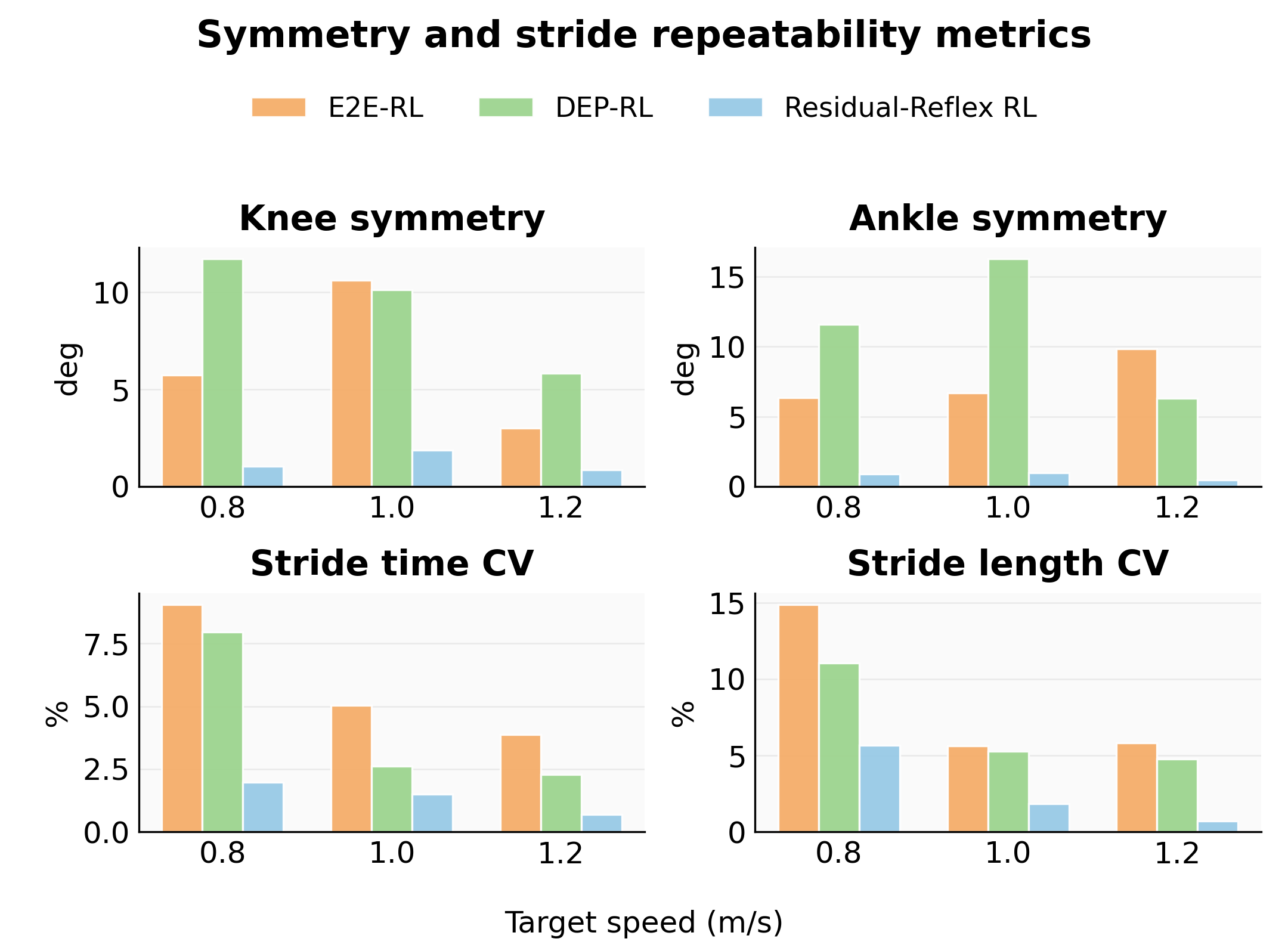}}
\caption{Left--right waveform differences and stride variability across target
speeds. Lower values indicate more symmetric and more repeatable gait cycles.}
\Description{Bar charts compare bilateral waveform differences and stride
variability. Residual-Reflex RL generally has the lowest values.}
\label{fig:symmetry_variability_bars}
\end{figure}

\subsubsection{Qualitative Motion Analysis}
\begin{figure}[!t]
\centering
\includegraphics[width=\columnwidth,height=0.72\textheight,keepaspectratio]{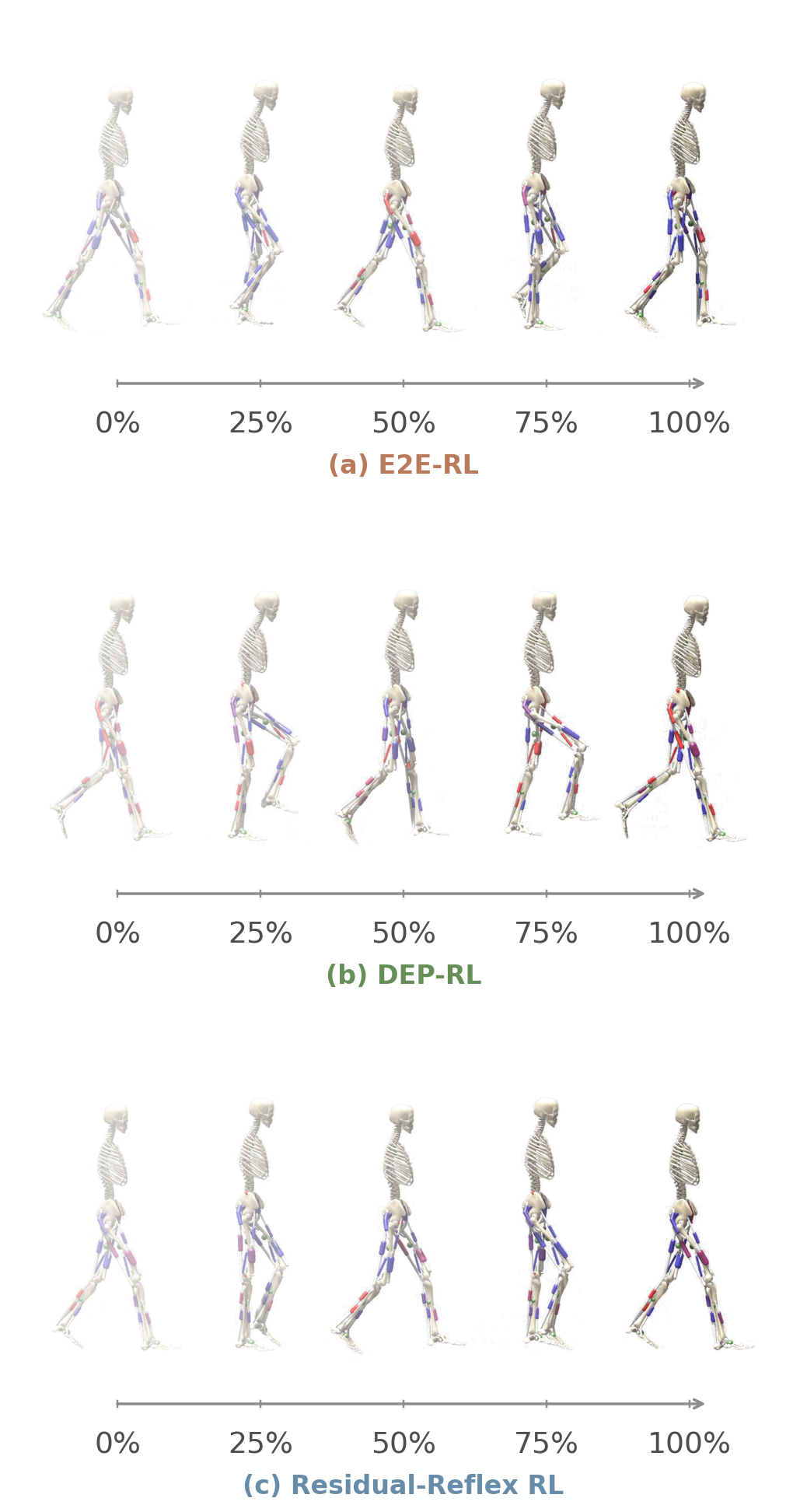}
\caption{Qualitative gait snapshots over one detected gait cycle at 1.2~m/s
for (a) E2E-RL, (b) DEP-RL, and (c) Residual-Reflex RL. Within each panel,
frames were sampled at 0\%, 25\%, 50\%, 75\%, and 100\% of the gait cycle
detected from the corresponding STO trajectory and synchronized video.}
\Description{Five gait phases for each controller show posture progression
through one walking cycle.}
\label{fig:qualitative_snapshots}
\end{figure}

The qualitative gait snapshots in Fig.~\ref{fig:qualitative_snapshots} provide a visual interpretation of the quantitative results presented in the preceding sections. Representative frames sampled at 0\%, 25\%, 50\%, 75\%, and 100\% of the gait cycle illustrate how the different controllers organize stance, push-off, and swing throughout a complete walking cycle.

Residual-Reflex RL produces the most natural-looking gait sequence. The body remains upright throughout the gait cycle, while the transition between stance and swing is smooth and well coordinated. These observations are consistent with its higher joint-kinematic, GRF, and gait-consistency scores, indicating that the proposed controller achieves both physiologically plausible joint motion and stable foot--ground interaction.

E2E-RL also generates sustained walking and maintains visually reasonable postures throughout the gait cycle. However, when interpreted together with the quantitative results, the motion exhibits less coordinated limb movement and less consistent weight transfer, explaining its lower GRF and gait-consistency scores despite relatively good joint-kinematic performance in some conditions.

DEP-RL exhibits the largest qualitative deviations. Around 25\% of the gait cycle, the body leans further forward while the swing leg adopts a less natural knee--ankle configuration. Similar deviations remain visible near 75\% of the gait cycle, resulting in less coordinated stance-to-swing transitions. These observations are consistent with its lower joint-kinematic, GRF, and gait-consistency scores.

The accompanying \suppvideo{} further illustrates these differences over continuous motion. Compared with the sampled snapshots, the video more clearly reveals the temporal characteristics of weight transfer, limb coordination, and stride-to-stride consistency throughout the gait cycle.

\subsection{Adaptation to Plantarflexor Weakness}
\label{sec:weakness-adaptation}
Beyond nominal walking, this experiment evaluates whether the trained Residual-Reflex RL controller adapts to changes in musculoskeletal capacity without retraining. Plantarflexor weakness was simulated by reducing the maximum isometric force of the soleus muscles while keeping the trained 1.2 m/s policy fixed. Soleus strength was varied from 83\% to 100\% of the nominal value. Residual-Reflex RL was compared with the CMA-ES-optimized reflex controller under the same weakened musculoskeletal models. To analyze the adaptation strategy, we additionally report the root-mean-square (RMS) magnitude of each residual-action channel,

\begin{equation}
A_i^{\mathrm{RMS}} =
\sqrt{\frac{1}{T}\sum_{t=1}^{T}a_i(t)^2},
\end{equation}
where $a_i(t)$ is the $i$th residual action and $T$ is the number of evaluated policy steps. The RMS magnitude quantifies how strongly each residual channel is utilized regardless of sign.

\subsubsection{Survival comparison}
The survival results immediately distinguish the two controllers. As summarized in Table~\ref{tab:weakness_survival}, Residual-Reflex RL completed the full 25 s evaluation at all tested weakness levels, whereas the CMA-ES-optimized reflex controller terminated early for every weakened model and remained stable only under the nominal condition. These results demonstrate that online residual-reflex modulation substantially improves robustness to changes in musculoskeletal capacity compared with fixed offline-optimized reflex parameters.

\subsubsection{Speed adaptation analysis}
\begin{figure}[!t]
\centerline{\includegraphics[width=0.78\columnwidth]{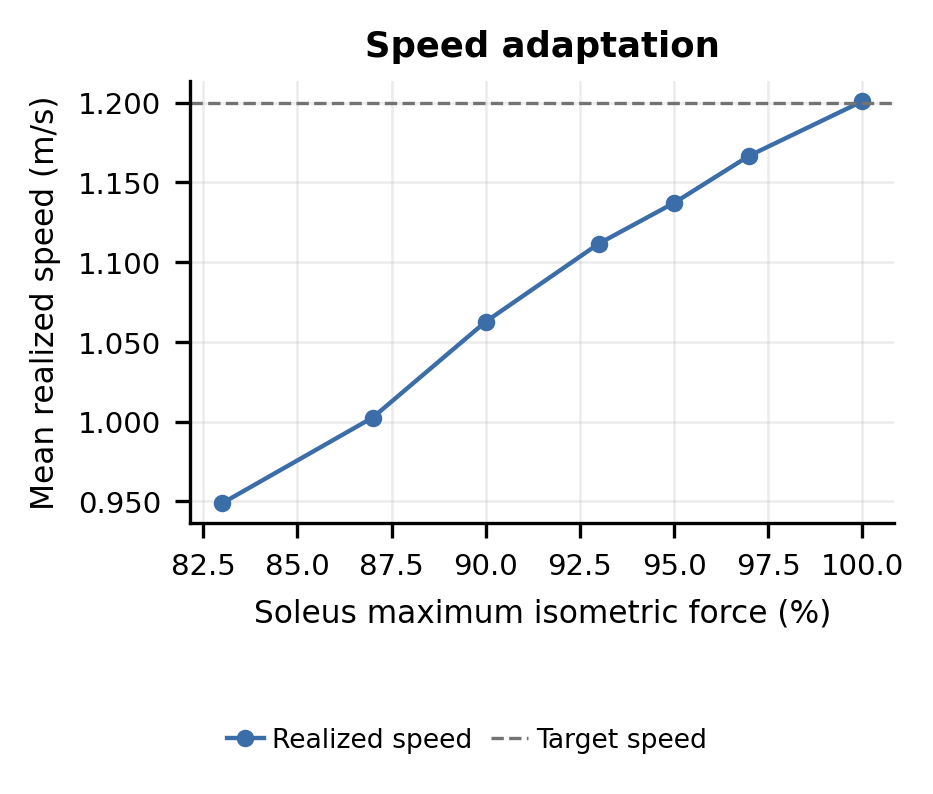}}
\caption{Speed adaptation under plantarflexor weakness. The same trained
Residual-Reflex RL policy remains stable across weakened soleus models but
adopts a slower realized speed as force capacity is reduced.}
\Description{Realized speed decreases smoothly as soleus maximum isometric
force is reduced from 100 to 83 percent.}
\label{fig:weakness_speed_adaptation}
\end{figure}

\begin{figure}[!t]
\centerline{\includegraphics[width=\columnwidth]{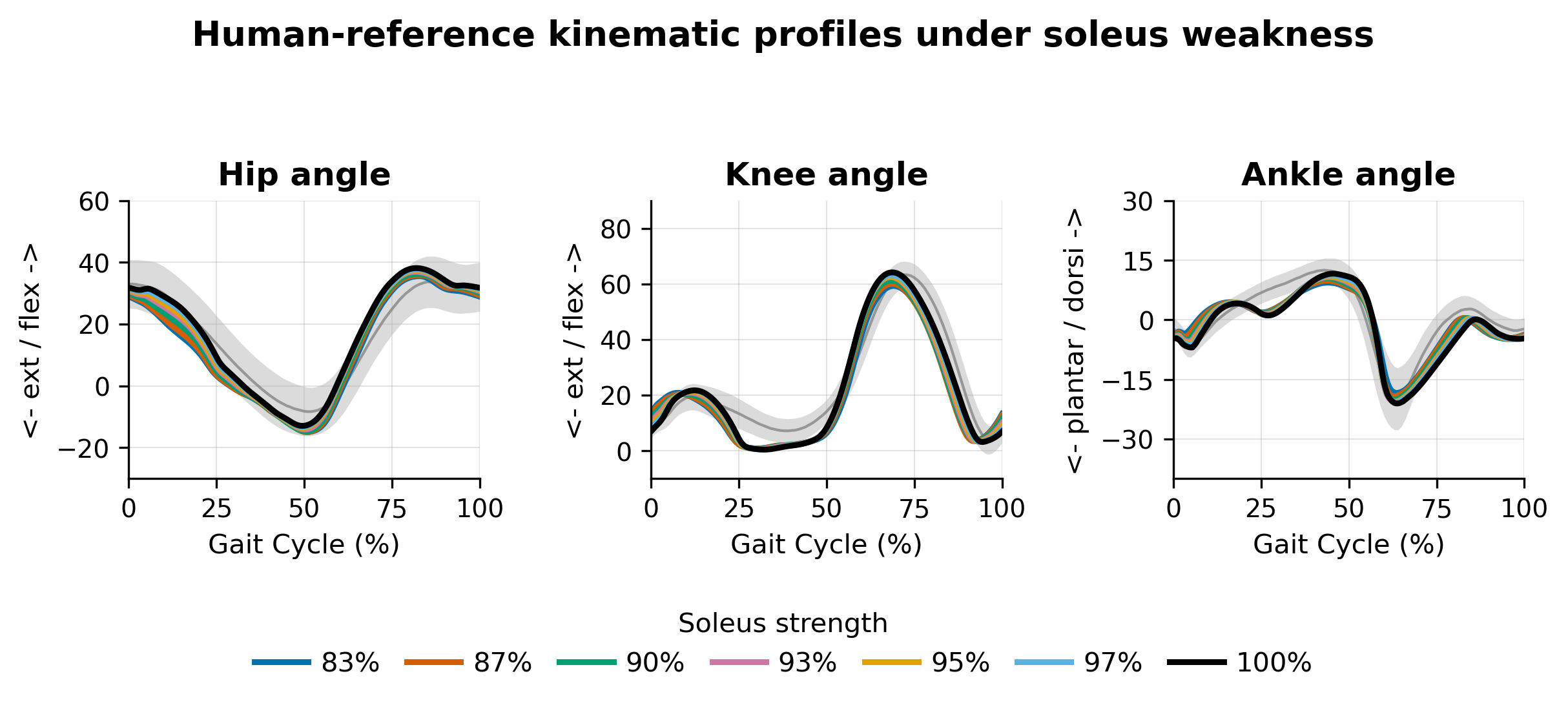}}
\caption{Human-reference kinematic profiles under soleus weakness. Colored
curves show the mean normalized gait-cycle profiles generated by the same
trained Residual-Reflex RL policy under different soleus strength levels; the
gray bands denote the human-reference ranges.}
\Description{Hip, knee, and ankle gait-cycle curves at seven soleus strength
levels remain near the human-reference ranges.}
\label{fig:weakness_kinematic_profiles}
\end{figure}

The speed adaptation results further reveal how the learned controller responds to reduced plantarflexor capacity. As shown in
Table~\ref{tab:weakness_adaptation} and
Fig.~\ref{fig:weakness_speed_adaptation}, the realized walking speed decreased gradually from 1.20 m/s under the nominal model to 0.95 m/s at 83\% soleus strength. Rather than maintaining the commanded speed at the expense of stability, the controller adopted a slower but stable gait as muscle capacity decreased. This behavior resembles a common compensatory strategy observed in human walking, where reduced plantarflexor strength is accompanied by a reduction in preferred walking speed. Importantly, the slower gait was not merely a degraded survival behavior. As shown in Fig.~\ref{fig:weakness_kinematic_profiles}, the generated joint trajectories remained close to the human-reference profiles across all tested weakness levels, yielding a joint-kinematic score of 89.0\% even at 83\% soleus strength. These results demonstrate that the proposed controller accommodates substantial changes in muscle capacity while preserving physiologically plausible locomotion without retraining.

The accompanying \suppvideo{} further illustrates the representative 83\% weakness condition. Residual-Reflex RL maintains a recognizable stance--swing sequence and upright progression, whereas the CMA-ES-optimized reflex controller loses balance and terminates. The synchronized gait-cycle curves further show the evolution of the hip, knee, ankle, and vertical GRF profiles during the two rollouts.

\subsubsection{Residual-action analysis}
The residual-action RMS values further explain the adaptation strategy adopted by the learned controller. Channel definitions are given in Table~\ref{tab:residual_actions}. Here we focus on how their modulation changes with soleus capacity. The soleus-related channel $a_3$ exhibits the largest RMS magnitude throughout the tested weakness range, indicating that ankle support and propulsion remain strongly regulated as plantarflexor capacity decreases. The policy does not compensate by monotonically increasing $a_3$. Instead, it accepts a lower walking speed while redistributing modulation across the remaining pathways. The stronger response of $a_4$ under greater weakness is consistent with increased reliance on knee-extension support when ankle push-off capacity is reduced, whereas $a_1$ and $a_2$ change more gradually. Together, the paired trends of ($a_1$,$a_2$) and ($a_3$,$a_4$) indicate coordinated redistribution across swing-related and stance-support pathways rather than single-parameter compensation. These results demonstrate that the four-dimensional residual action provides a compact adaptive interface for online neuromuscular regulation, enabling the same trained policy to generalize across weakened musculoskeletal models without retraining.

\begin{table}[!t]
\centering
\caption{Survival comparison under soleus weakness. Full episode duration is
25~s. The CMA-ES-optimized reflex controller remains stable only in the nominal
100\% model, whereas Residual-Reflex RL completes the episode across all tested
weakness levels without retraining.}
\label{tab:weakness_survival}
\small
\setlength{\tabcolsep}{1.8pt}
\renewcommand{\arraystretch}{1.1}
\begin{tabular*}{\columnwidth}{@{\extracolsep{\fill}}c c c c c}
\toprule
\textbf{Soleus} &
\makecell{\textbf{Residual}\\\textbf{duration}\\\textbf{(s)}} &
\makecell{\textbf{Residual}\\\textbf{full}} &
\makecell{\textbf{CMA-ES}\\\textbf{duration}\\\textbf{(s)}} &
\makecell{\textbf{CMA-ES}\\\textbf{full}} \\
\midrule
83\%  & 25.0 & Yes & 1.27 & No \\
87\%  & 25.0 & Yes & 1.28 & No \\
90\%  & 25.0 & Yes & 1.28 & No \\
93\%  & 25.0 & Yes & 1.29 & No \\
95\%  & 25.0 & Yes & 1.48 & No \\
97\%  & 25.0 & Yes & 1.47 & No \\
100\% & 25.0 & Yes & 25.0 & Yes \\
\bottomrule
\end{tabular*}
\end{table}

\begin{table}[!t]
\centering
\caption{Residual-Reflex RL under soleus weakness. The target speed was
1.2~m/s. All listed residual rollouts reached 1000 steps. Channel $a_3$ is the
soleus force-feedback modulation defined in Table~\ref{tab:residual_actions}.}
\label{tab:weakness_adaptation}
\small
\setlength{\tabcolsep}{4pt}
\renewcommand{\arraystretch}{1.1}
\begin{tabular*}{\columnwidth}{@{\extracolsep{\fill}}c c c c}
\toprule
\makecell{\textbf{Soleus}\\\textbf{strength}} &
\makecell{\textbf{Realized speed}\\\textbf{(m/s)}} &
\makecell{\textbf{Speed error}\\\textbf{(m/s)}} &
\textbf{RMS $\boldsymbol{a_3}$} \\
\midrule
83\%  & 0.949 & 0.251 & \textbf{0.846} \\
87\%  & 1.002 & 0.198 & \textbf{0.851} \\
90\%  & 1.063 & 0.137 & \textbf{0.878} \\
93\%  & 1.112 & 0.088 & \textbf{0.885} \\
95\%  & 1.137 & 0.063 & \textbf{0.889} \\
97\%  & 1.167 & 0.033 & \textbf{0.895} \\
100\% & 1.201 & 0.001 & \textbf{0.898} \\
\bottomrule
\end{tabular*}
\end{table}

\begin{figure}[!t]
\centerline{\includegraphics[width=0.78\columnwidth]{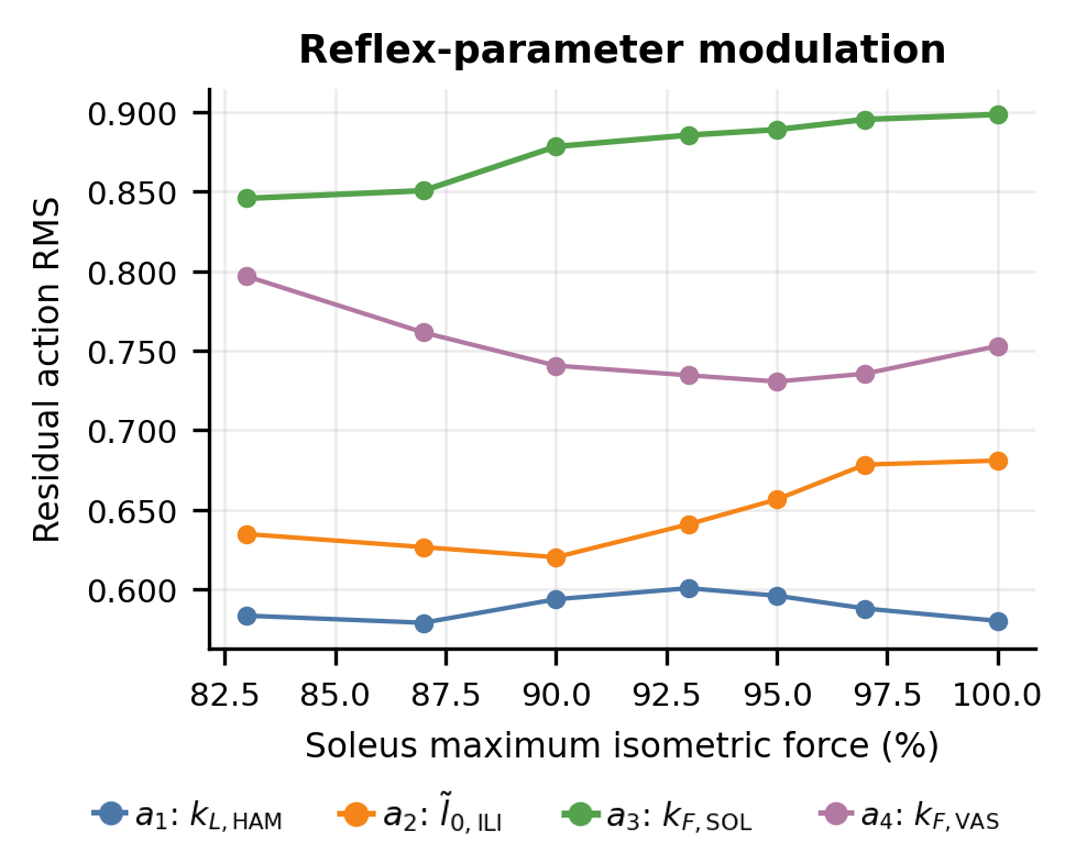}}
\caption{Residual-action modulation under plantarflexor weakness. RMS values
show that the trained policy continues to adjust all four selected reflex
channels. Channel $a_3$ modulates soleus force feedback for ankle
support/propulsion, whereas $a_4$ modulates vasti force feedback for knee
extension and stance support.}
\Description{Four curves show the root-mean-square magnitude of each residual
reflex action across soleus strength levels. The soleus-related third channel
has the largest magnitude.}
\label{fig:weakness_residual_action_rms}
\end{figure}

\subsection{Recovery from External Pushes}
\label{sec:push-recovery}
This experiment evaluates whether the trained Residual-Reflex RL controller can recover from external disturbances without retraining. The same 1.2 m/s policy used in the nominal walking and plantarflexor-weakness experiments was retained without further learning. Backward force pulses were applied to the torso for 0.1 s starting at 3.0 s. Perturbation magnitudes of 75 N and 100 N were evaluated under both single-push and repeated-push (every 5 s) conditions. Residual-Reflex RL was compared with the CMA-ES-optimized reflex controller under identical perturbation settings.

As summarized in Table~\ref{tab:push_perturbation}, Residual-Reflex RL successfully recovered from both 75 N and 100 N single backward pushes and maintained walking for the full 25 s evaluation. Under repeated perturbations, the controller also completed the full evaluation under the 75 N condition, whereas the more challenging 100 N repeated pushes resulted in termination after 20.1 s. In contrast, the CMA-ES-optimized reflex controller consistently failed after approximately 4.5--4.6 s under all perturbation conditions. These results demonstrate that the learned residual-reflex policy substantially improves recovery capability beyond fixed offline-optimized reflex gains.

\begin{figure*}[!t]
\centering
\includegraphics[width=0.94\textwidth]{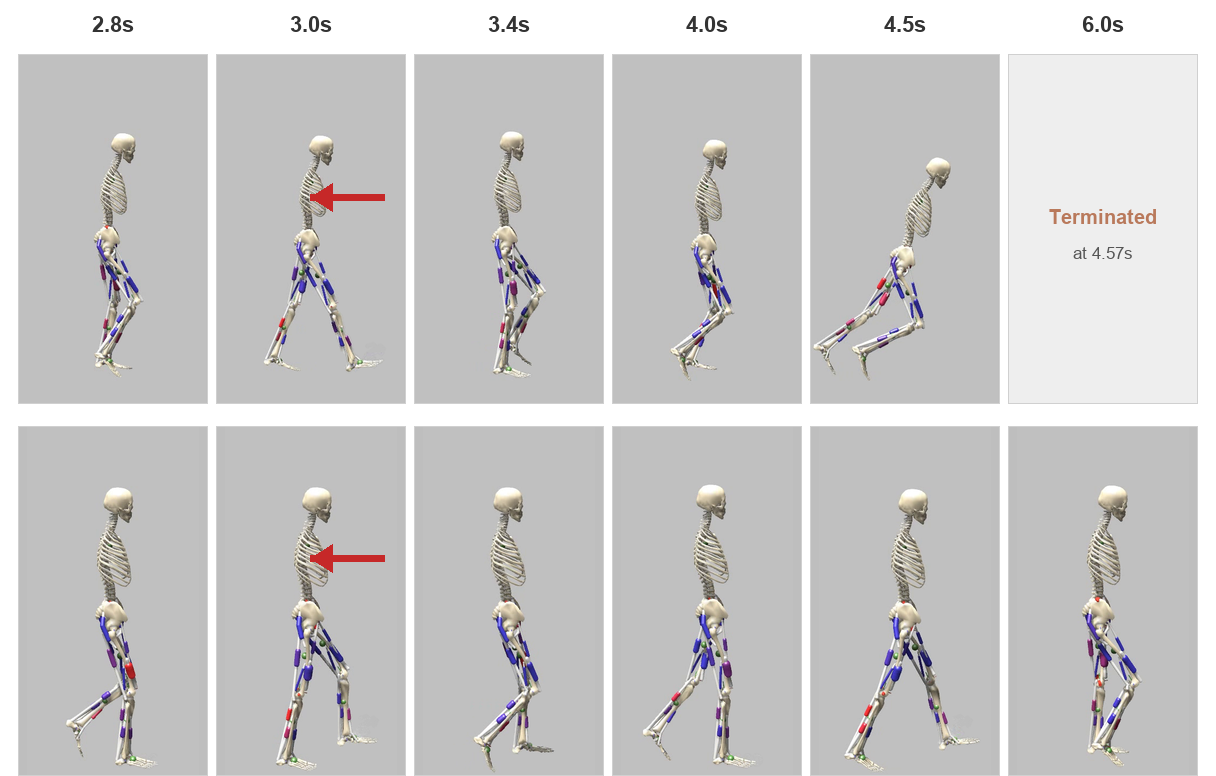}
\caption{Time-aligned response to a 100~N backward torso push applied for 0.1~s
at $t=3.0~\mathrm{s}$. Columns show synchronized frames from 2.8~s to 6.0~s. The
CMA-ES-optimized reflex controller (top) progressively loses upright balance
and terminates at 4.57~s. Residual-Reflex RL (bottom) modifies the following
steps and continues walking for the full 25~s rollout under the same
perturbation. Red arrows indicate the applied force.}
\Description{Two synchronized frame sequences compare recovery after a
backward torso push. The optimized reflex controller falls, whereas
Residual-Reflex RL reorganizes its steps and continues walking.}
\label{fig:push_snapshots}
\end{figure*}

Figure~\ref{fig:push_snapshots} provides a qualitative comparison of the recovery process under the representative 100 N single-push condition. Before the perturbation, both controllers produce stable walking. Following the force pulse, the CMA-ES-optimized reflex controller progressively loses balance and terminates, whereas Residual-Reflex RL reorganizes the subsequent steps, restores upright walking, and completes the full evaluation. The matched frames therefore distinguish sustained recovery from trajectories that merely delay failure.

The accompanying \suppvideo{} first illustrates the single-push recovery process by comparing Residual-Reflex RL under 75~N and 100~N backward pushes with the CMA-ES-optimized reflex controller under a 100~N backward push. Residual-Reflex RL reorganizes foot placement and resumes periodic walking after the disturbance, whereas the CMA-ES-optimized reflex controller fails to regain a stable gait. The video then compares Residual-Reflex RL under repeated 75~N and 100~N backward pushes applied every 5~s. The controller remains stable throughout the 75~N trial but terminates after 20.1~s under repeated 100~N pushes, illustrating the boundary of its recovery capability. Together with the plantarflexor-weakness experiments, these results demonstrate that residual-reflex modulation provides a flexible online adaptation mechanism for changes in both musculoskeletal capacity and external disturbances.

\begin{table}[!t]
\centering
\caption{External backward push perturbation results. Full episode duration is
25~s.}
\label{tab:push_perturbation}
\small
\setlength{\tabcolsep}{1.8pt}
\renewcommand{\arraystretch}{1.08}
\begin{tabular*}{\columnwidth}{@{\extracolsep{\fill}}c c c c c}
\toprule
\textbf{Method} & \textbf{Perturbation} & \makecell{\textbf{Force}\\\textbf{(N)}} &
\makecell{\textbf{Duration}\\\textbf{(s)}} &
\makecell{\textbf{Speed}\\\textbf{(m/s)}} \\
\midrule
\multirow[c]{4}{*}{\makecell[c]{Residual-\\Reflex RL}} &
\multirow{2}{*}{Single push} & 75  & 25.0 & 1.188 \\
& & 100 & 25.0 & 1.176 \\
\cmidrule(lr){2-5}
& \multirow{2}{*}{Every 5~s} & 75  & 25.0 & 1.129 \\
& & 100 & 20.1 & 1.091 \\
\midrule
\multirow[c]{4}{*}{\makecell[c]{CMA-ES-\\optimized\\Reflex}} &
\multirow{2}{*}{Single push} & 75  & 4.5 & 1.098 \\
& & 100 & 4.6 & 1.069 \\
\cmidrule(lr){2-5}
& \multirow{2}{*}{Every 5~s} & 75  & 4.5 & 1.098 \\
& & 100 & 4.6 & 1.069 \\
\bottomrule
\end{tabular*}
\end{table}

\section{Discussion}
\label{sec:discussion}
The experimental results demonstrate that regulating an existing neuromuscular controller through reinforcement learning provides a practical way to improve both physiological plausibility and adaptability in muscle-driven locomotion. Under nominal walking conditions, Residual-Reflex RL consistently achieved better joint kinematics, more human-like ground reaction forces, and higher gait consistency than end-to-end muscle-control policies. The weakness and perturbation experiments further showed that the same trained policy remained effective under changes in musculoskeletal capacity and external disturbances without retraining. Together, these results indicate that reinforcement learning can benefit from regulating neuromuscular control mechanisms rather than directly generating muscle actions.

The observed improvements in joint kinematics, GRF morphology, and gait consistency can be understood from the learning formulation itself. In end-to-end muscle control, reinforcement learning directly operates in a high-dimensional and redundant muscle action space, where both muscle activation and the functional relationships among muscles must be learned from task rewards. In the proposed framework, the underlying neuromuscular topology is fixed by the phase-dependent reflex controller, such that the functional organization of sensory feedback pathways is explicitly defined. Reinforcement learning therefore no longer needs to construct this neuromuscular organization from scratch, but instead learns how to regulate a small set of biomechanically meaningful reflex parameters according to the current state. This allows learning to focus on state-dependent adaptation while preserving the existing neuromuscular organization.

The weakness and perturbation experiments further clarify why this formulation improves adaptability. The CMA-ES-optimized reflex controller also relies on physiological feedback, but its parameters remain fixed once optimization is completed. In contrast, Residual-Reflex RL preserves the same neuromuscular control structure while continuously adjusting the selected reflex parameters according to the current musculoskeletal state. This adaptation mechanism becomes particularly important when muscle capacity changes or external disturbances occur, because the controller can continuously regulate the underlying reflex parameters instead of relying on a single fixed parameter set. These findings demonstrate that reflex-informed neuromuscular regulation provides an effective formulation for improving both physiological plausibility and adaptability in muscle-driven locomotion.

\section{Conclusion}
We presented a Reflex-Informed Neuromuscular Reinforcement Learning framework for adaptive muscle-driven locomotion. Instead of directly generating muscle excitations, the proposed framework learns to regulate four biomechanically meaningful residual parameters of a phase-dependent reflex controller. This formulation preserves the underlying neuromuscular control structure while enabling reinforcement learning to adapt reflex behavior online through a compact and interpretable control interface.

Experiments in a Hyfydy-based musculoskeletal simulation demonstrated that the proposed framework consistently improved physiological plausibility under nominal walking conditions. Compared with end-to-end muscle-control policies, Residual-Reflex RL generated more human-like joint kinematics, ground reaction forces, and gait consistency across walking speeds of 0.8, 1.0, and 1.2~m/s. Beyond nominal walking, the same trained policy remained effective under plantarflexor weakness and external perturbations without retraining, demonstrating improved adaptability while preserving physiologically plausible locomotion.

These results demonstrate that integrating reinforcement learning with an existing neuromuscular control structure provides an effective formulation for simultaneously improving physiological plausibility and adaptability in muscle-driven locomotion. The proposed framework offers a compact and biologically meaningful learning interface that is applicable to muscle-driven character animation, computational studies of human locomotion, and bio-inspired locomotor control. Future work will extend the framework to continuous speed transitions, broader musculoskeletal morphologies, and more diverse physical interactions.

\FloatBarrier
\bibliographystyle{plainnat}
\bibliography{references}

\end{document}